\documentclass[journal]{IEEEtran}
\usepackage{cite,amsmath,amssymb,graphicx,tabularx,multirow,booktabs,caption,xurl}
\begin{document}
\title{R2D-RL: A RoboCup 2D Soccer Environment for Multi-Agent Reinforcement Learning}
\author{Haobin Qin, Baofeng Zhang, Hidehisa Akiyama, and Keisuke Fujii,~\IEEEmembership{Member,~IEEE}%
\thanks{Haobin Qin, Baofeng Zhang, and Keisuke Fujii are with the Graduate School of Informatics, Nagoya University, Furo-cho, Chikusa, Nagoya, Aichi, Japan.}%
\thanks{Hidehisa Akiyama is with the School of Information and Data Sciences, Nagasaki University, Nagasaki, Japan.}%
\thanks{Corresponding author: Keisuke Fujii (email: fujii@i.nagoya-u.ac.jp).}%
\thanks{This work was financially supported by JST SPRING, Grant Number JPMJSP2125; NEDO Intensive Support Program for Young Promising Researchers, Grant Number 24021654; JSPS KAKENHI Grant Number 23H03282.}}
\maketitle
\begin{abstract}
Robot soccer is a challenging testbed for multi-agent reinforcement learning because it combines partial observability, cooperative and adversarial interaction, sparse rewards, and long-horizon tactical behavior. RoboCup 2D Soccer Simulation (RCSS2D) provides a mature robot-soccer platform, but its competition-oriented server-client architecture is difficult to use directly with modern Python-based MARL workflows. We introduce R2D-RL, a reinforcement learning environment that connects RCSS2D and HELIOS-based player clients to a Python MARL interface through shared-memory communication and cycle-level synchronization. {R2D-RL supports both full-field and scenario-based training, with opponents controlled by HELIOS-base, Base and Hybrid action interfaces, action masking, an auxiliary MaxEPV reward based on improvements in expected possession value (EPV), and parallelized execution.} We provide front-goal scenarios and a full-field benchmark, together with baseline results. {With 12 parallel 11-vs-11 environments on a single CPU socket, sampling throughput reaches 427.48 and 359.49 joint environment steps/s for the Base and Hybrid interfaces, respectively, while end-to-end training throughput ranges from 132.83 to 330.09 joint steps/s. After 30 million training steps, MAPPO and IPPO achieve mean goal differences of \(-12.83\) and \(-12.71\) against HELIOS-base, respectively, highlighting the remaining difficulty of full-match learning.}
\end{abstract}

\begin{IEEEkeywords}
Multi-agent reinforcement learning, RoboCup 2D Soccer Simulation, robot soccer, action masks, reinforcement learning environment.
\end{IEEEkeywords}

\section{Introduction}\label{sec:introduction}

Soccer is a representative domain for learning-based sports analytics, where strategic evaluation and next-play decision making require reasoning about many interacting players over space and time \cite{fujii2025machinelearning}. It combines partial observability, simultaneous cooperative and adversarial interaction, sparse scoring events, and long-horizon tactical organization. These properties make football a natural domain for studying multi-agent reinforcement learning and tactical decision making.

To support football reinforcement learning and multi-agent decision research, different types of football simulation and learning platforms have been proposed. One line of work is oriented toward football decision making and MARL benchmarks, where football is organized as a learning problem for algorithm evaluation, reward design, scenario-based learning, or large-scale training; representative examples include Google Research Football and NFootball \cite{kurach2020gfootball,fujii2024adaptive}. Another line of work is oriented toward robot soccer and embodied control, where football interaction is coupled with physical movement, low-level control, simulator protocols, and competition-style multi-agent interaction, with RoboCup being the most representative example.

RoboCup 2D Soccer Simulation (RCSS2D) \cite{noda1996soccer,rcssserver_site} represents a long-standing platform in this control-oriented robot-soccer tradition. It provides a full-field match setting with game rules, referee control, physics, stamina, noisy sensing, and a distributed server-client architecture. Around this simulator, the RoboCup 2D community has accumulated mature base teams and tactical behavior modules. In particular, HELIOS-based teams provide rule-based soccer behaviors such as passing, dribbling, shooting, interception, positioning, and set plays.

{Existing work improves either client development through Python or language-agnostic frameworks, such as Pyrus and the Cross Language Soccer Framework, or learning on restricted RCSS2D subtasks, such as Keepaway and Half Field Offense (HFO) \cite{zare2023pyrus,zare2024cross,stone2005keepaway,stone2006keepaway,kalyanakrishnan2006half}.}

{As reinforcement learning and multi-agent learning have developed, football simulation platforms are increasingly expected to support match execution together with learning-oriented environment functions, including standardized step-based interaction, reproducible task control, training signals, parallel sampling, and benchmark settings.} Google Research Football has attracted broad attention in football reinforcement-learning research in recent years precisely because it matches these requirements. By contrast, although RCSS2D provides full-field match mechanics and the HELIOS behavior ecosystem, it lacks a synchronization interface for modern MARL training, making it difficult to use directly for full-field reinforcement-learning research.

{This paper presents R2D-RL, a Python MARL environment for RoboCup 2D that integrates the original RCSS2D server and modified HELIOS clients with a Python learning interface through shared-memory communication and cycle-level synchronization. It supports both full-field and scenario-based training, with HELIOS-base-controlled opponents, Base and Hybrid action interfaces, controllable agents, action masking, and parallel execution.}

The main contributions are as follows:
\begin{enumerate}
\item We design a shared-memory communication and cycle-level synchronization architecture that connects the original RCSS2D server-client workflow with Python MARL training and organizes it as step-synchronous reinforcement-learning interactions.
\item We provide a Python MARL interface that exposes local observations, centralized state, action masks, rewards, termination signals, and reset operations for learning algorithms.
\item We construct an R2D-RL benchmark suite covering front-goal scenarios and 11-vs-11 full-field learning.
\item We define two policy-level action spaces: a parameterized Hybrid interface over selected RCSS2D body commands and a HELIOS-derived Base discrete action space with corresponding action masks.
\end{enumerate}

\begin{figure}[t]
  \centering
  \includegraphics[width=0.85\columnwidth]{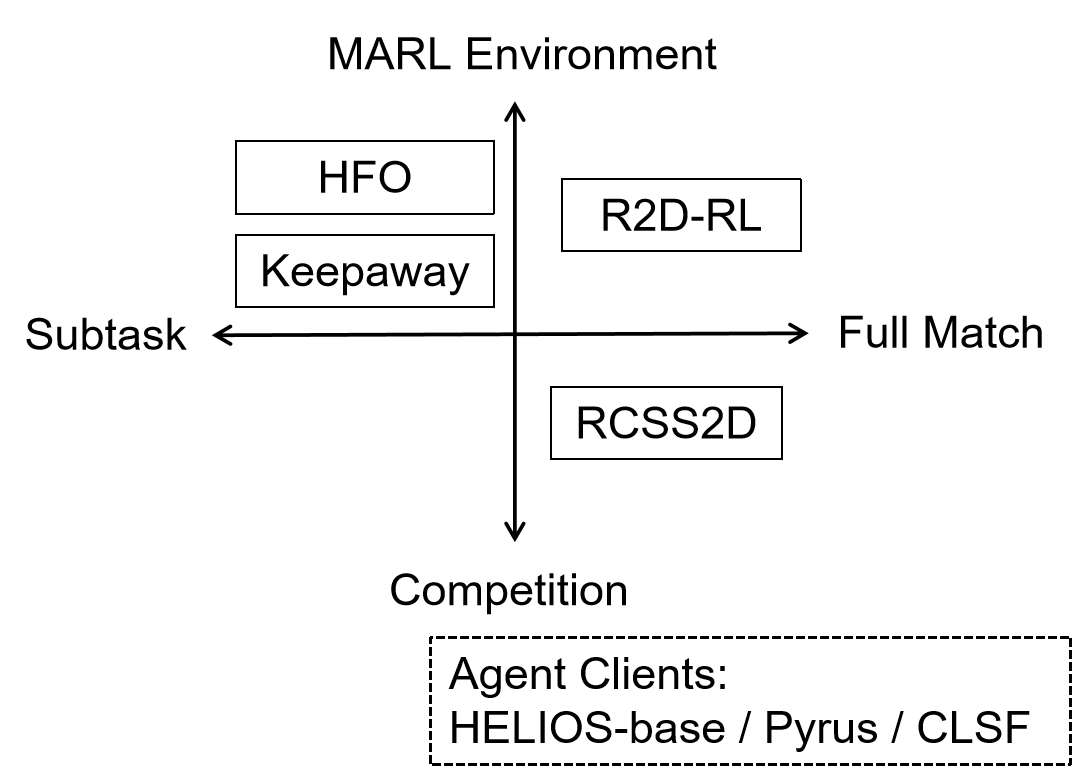}
  \caption{Positioning of R2D-RL in the RoboCup 2D ecosystem.}
  \label{fig:ecosystem_positioning}
\end{figure}
\section{Related Work}\label{sec:related_work}

\subsection{{Co-evolution of Football Learning Tasks and Algorithmic Capabilities}}

RoboCup is a long-running research initiative in robotics and artificial intelligence, originally proposed by Kitano et al. as a challenge problem for intelligent agents in a dynamic, real-time, multi-agent domain \cite{kitano1997robocup}. Within RoboCupSoccer, the Simulation League is one of the oldest branches and is focused on artificial intelligence and team strategy \cite{robocup_simulation_league}. RoboCup 2D Soccer Simulation (RCSS2D) is a classic platform in this line of work. In the 2D Simulation League, two teams of eleven autonomous software agents play soccer in a two-dimensional virtual stadium represented by a central SoccerServer \cite{robocup_2d_league,noda1996soccer,rcssserver_site}. The server maintains the players, ball, physics, and match state; player clients issue commands from local perceptions; the coach provides global match information; and the trainer supports reset and scenario control. {This distributed architecture provides a rich setting for multi-agent cooperation and competition. However, RCSS2D was originally designed for robotic football competition and agent development, predating modern conventions for reinforcement-learning environments.}

Full-field RCSS2D games involve long horizons, sparse scoring events, and many simultaneously acting agents. Against this background, several lightweight RCSS2D-derived subtask environments have been introduced to support learning-algorithm development and testing. Keepaway simplifies football into a small-sided ball-possession task and has been used as a testbed for algorithms studying high-level decisions such as holding the ball, passing, and maintaining space under multi-agent interaction \cite{stone2005keepaway,stone2006keepaway}. Half Field Offense (HFO) restricts the problem to a half-field attacking environment with clearer state, action, and reward interfaces for testing algorithms for possession, passing, shooting, attacking cooperation, and multi-agent reinforcement learning \cite{kalyanakrishnan2006half}.

{The progression from Keepaway to HFO shows that subtasks reduced the physical scope of the game and aligned the learning problem with the algorithmic capabilities available at the time.} {Methods used in Keepaway could address a restricted set of high-level possession decisions, whereas the HFO study reported that methods successful in Keepaway did not directly scale to a half-field attacking task with sparser rewards, larger state and action spaces, and more complex coordination \cite{stone2005keepaway,kalyanakrishnan2006half}.} {Reducing the number of players, field size, and task objective consequently provided more tractable conditions for training and analysis, but omitted full-field spatial organization, long-horizon attack--defense transitions, and team decision making under complete match rules.}

{Subsequent advances in multi-agent reinforcement learning increased the ability to address larger cooperative problems.} {Centralized training with decentralized execution (CTDE) allows learning algorithms to use additional global information during training while retaining local decision making at execution time.} {MADDPG uses centralized critics to mitigate the non-stationarity encountered by independent learners, whereas QMIX, a monotonic value-function factorization method, trains decentralized policies from global state information \cite{lowe2017multi,rashid2018qmix}.} {PPO-based multi-agent methods have subsequently also demonstrated strong training performance across several cooperative benchmarks \cite{yu2022surprising}.} {Football retains long-horizon exploration, credit-assignment, and opponent non-stationarity challenges. These developments provide an algorithmic basis for expanding from localized tasks with few agents toward more complete matches.}

{The introduction of Google Research Football (GRF) can be viewed as an important milestone in this progression.} {As deep reinforcement learning, parallel sampling, and standardized environment interfaces matured, GRF organized full football matches as a reproducible reinforcement-learning benchmark, making full-field football increasingly amenable to systematic training and algorithm comparison \cite{kurach2020gfootball}.} {At the same time, GRF retains a broad set of localized Football Academy scenarios, indicating that improved algorithms have not eliminated the sparse-reward and coordination difficulties of complete matches.} {Football learning environments and learning algorithms have therefore co-evolved: environment simplification made localized football problems accessible to earlier methods, while advances in algorithms, sampling systems, and environment interfaces enabled learning tasks to expand toward complete matches.}

\subsection{{Different Research Orientations in Football Learning Environments}}

{Beyond GRF and the RoboCup 2D ecosystem, football learning environments organize the problem according to different research objectives, including task scope, simulator complexity, action abstraction, and data source.}

Unity ML-Agents soccer tasks, including Soccer Twos and Strikers vs. Goalie, use compact arenas, ray-cast observations, explicit rewards, and discrete branched actions that mainly correspond to movement and rotation, making them lightweight benchmarks for competitive and cooperative MARL \cite{juliani2018unity,unity_ml_agents_examples}. {This task-level simplification reduces learning and experimentation costs, but does not aim to reproduce a complete football match workflow.}

Recent simplified 11-vs-11 simulated robotic football work builds a custom 2D simulator and uses a simpler continuous action representation to improve training throughput for full-team MARL \cite{smit2023scaling}. {This study combines attention, parameter sharing, and opponent-stabilization techniques on the algorithmic side with a high-throughput simulator that simplifies player interactions, boundary handling, and parts of the match rules.} {The design retains the complete multi-agent scale while trading simulator complexity for sampling efficiency and training tractability, but does not model every physical process and match mechanism of a complete competition simulator.}

NFootball and the adaptive action-supervision framework of Fujii et al. connect football decision making with professional tracking data by using real-world multi-agent demonstrations and simulation-based tasks to study action supervision in reinforcement learning \cite{fujii2024adaptive}. {Unlike environments primarily designed for algorithm evaluation or training tractability, this work focuses on real-to-sim learning under a dynamics gap between real match data and the target simulator.} {Its objective is to balance the reproduction of observed behavior with policy generalization to new states, making dynamics mismatch and behavioral-distribution shift central challenges.}

\begin{table*}[t]
\caption{Comparison with representative football learning environments.}
\label{tab:football_env_comparison}
\centering
\scriptsize
\setlength{\tabcolsep}{3pt}
\renewcommand{\arraystretch}{1.08}
\begin{tabular}{@{}lllllll@{}}
\toprule
Environment & Scope & Simulator Base & Built-in AI & Action space & Action mask & Initial state \\
\midrule
GRF \cite{kurach2020gfootball} & Full-field+scenario & Football Engine & Yes & High-level & No & Position \\
Unity Soccer \cite{juliani2018unity,unity_ml_agents_examples} & Scenario & Unity & No & Low-level & No & Preset layout \\
Custom 2D Football \cite{smit2023scaling} & Full-field & Custom 2D & No & Low-level & No & Position \\
NFootball \cite{fujii2024adaptive} & Scenario & Python 2D & No & High-level & No & Tracking data \\
\textbf{R2D-RL (ours)} & \textbf{Full-field+scenario} & \textbf{RCSS2D} & \textbf{Yes} & \textbf{High-level discrete + param. hybrid} & \textbf{Yes} & \textbf{Pos.+dir.+vel.} \\
\bottomrule
\end{tabular}
\end{table*}

{Table~\ref{tab:football_env_comparison} summarizes how these environments support different research objectives through localized tasks, simulator simplification, or real tracking data, together with their action abstractions and initial-state controls.}

\subsection{{From the C++ RCSS2D Competition Ecosystem to a Python MARL Environment}}

{RCSS2D combines a football simulator with a distributed robotic-football system that has been continuously developed through long-running competitions. It provides complete match rules and referee procedures, decentralized execution through independent clients, noisy local perception, stamina and body-command models, and a mature behavior ecosystem built around established base teams such as HELIOS-base. RCSS2D therefore enables policy training and evaluation within an existing robotic-football competition ecosystem, rather than only within football tasks designed specifically as reinforcement-learning benchmarks.}

{The RCSS2D server, commonly used clients, base teams, and behavior libraries have traditionally been implemented primarily in C++.} The RoboCup 2D community has also accumulated strong hand-coded client teams, base code, and client-side development frameworks. HELIOS teams have won the RoboCup 2D Simulation League multiple times \cite{akiyama2013helios}. HELIOS-base is a widely used open-source base team in the community \cite{akiyama2014heliosbase,helios_base_repo}. It provides rule-based player-client behaviors such as passing, shooting, dribbling, interception, positioning, and set plays, and these behaviors can serve as action abstractions for learning environments.

{Modern deep-learning and reinforcement-learning research, by contrast, increasingly relies on Python-based training frameworks, experiment toolchains, and standardized environment interfaces. This creates a need to bridge the established C++ RCSS2D competition ecosystem with modern reinforcement-learning workflows. The shift extends beyond programming-language choice: the research workflow has moved from developing behaviors for individual competition clients toward organizing large-scale training experiments around standardized observation, action, reward, step, and reset interfaces.}

To make RoboCup 2D client development easier, Pyrus reimplements the RCSS2D base-code interface in Python, targeting researchers who want to prototype player-client behavior and integrate machine-learning methods without working directly with C++ base teams \cite{zare2023pyrus}. The Cross Language Soccer Framework takes a different route toward easier client development by retaining the high-performance HELIOS base code while exposing a gRPC-based, language-agnostic client architecture that supports player-client development in languages such as C\#, JavaScript, and Python \cite{zare2024cross}. {These frameworks reduce the language and engineering barriers to accessing RCSS2D client functionality and established football behaviors, making it easier to embed a learning model within an individual player client.}

Multi-agent reinforcement learning (MARL) usually models an environment as a joint interaction process. At each environment step, the environment provides observations to multiple agents, the agents produce a joint action according to their policies, and the environment executes the joint action, transitions to the next state, and returns rewards, termination signals, and new observations. For football tasks, this formulation is particularly important because the match state is jointly driven by the simultaneous decisions of multiple players.

This formulation highlights the distinction between client-side simulator access and a training-time MARL environment. Client-side frameworks facilitate player-client implementation and simulator communication. However, client-side access alone does not define synchronized environment steps or expose the training-time semantics required by MARL algorithms, including centralized state, rewards, action masks, reset control, and termination signals. Therefore, adapting RCSS2D for MARL training requires an environment-side layer in addition to player-client development support.

Adapting RCSS2D to this MARL formulation first requires organizing the information needed for centralized training with decentralized execution (CTDE) \cite{lowe2017multi,rashid2018qmix}. RCSS2D originally adopts a distributed competition structure in which each player is an independent client that sends commands to the server based on its own local perception. This structure is naturally aligned with decentralized execution. However, centralized training additionally requires the environment side to organize centralized state, local observations, joint actions, available actions, rewards, and termination signals. Thus, an RCSS2D-based MARL environment must preserve independent player-side execution while providing training-time global state and feedback.

The second issue is consistency between actions and state transitions. The original RCSS2D workflow is a real-time server loop: player clients must send commands to the server within the time window of each simulator cycle. If a command does not arrive within the corresponding cycle, the server still advances the match, and the command is not incorporated into the intended state transition. For reinforcement-learning training, the joint action submitted at an environment step must correspond to the subsequent transition, reward, and termination signal. Therefore, the MARL step, player commands, and server simulation cycle must be synchronized so that action execution and environment feedback belong to the same transition process.

The third issue is exposing the resulting process as a MARL environment interface. After training information and step synchronization are organized, the RCSS2D match workflow still needs to be wrapped as operations that learning frameworks can call, such as \texttt{reset()}, \texttt{step()}, observations, centralized state, available actions, rewards, and termination signals. This interface layer is what turns a competition-oriented server-client system into a training environment that can be used directly by modern MARL frameworks.

{Together, these requirements define the environment-layer gap addressed by R2D-RL: a cycle-synchronized Python MARL interface that retains the distributed RCSS2D clients and HELIOS-base behavior ecosystem.} {Figure~\ref{fig:ecosystem_positioning} summarizes this positioning across the RCSS2D competition platform, subtask environments, client frameworks, and a full-match MARL environment.}

\section{Methodology}\label{sec:methodology}

\subsection{System Overview}

\begin{figure*}[t]
\centering
\includegraphics[width=\textwidth]{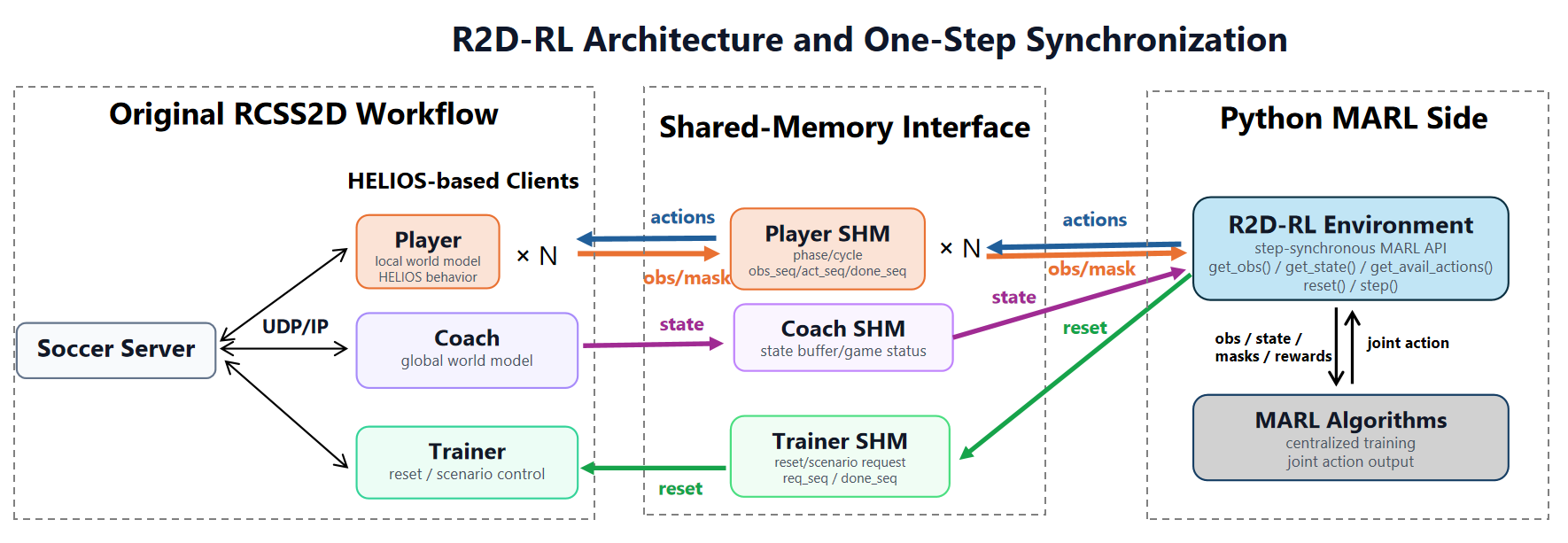}
\caption{R2D-RL architecture and one-step synchronization. The original RCSS2D workflow with the soccer server and modified HELIOS-based player, coach, and trainer clients is preserved. Player, coach, and trainer shared-memory buffers exchange actions, observations and action masks, centralized state, and reset requests with the Python MARL environment, which provides step-synchronous interaction for MARL algorithms.}
\label{fig:architecture}
\end{figure*}

R2D-RL is implemented as a Python MARL layer around the original RCSS2D server-client workflow. It runs the soccer server together with modified HELIOS player clients, coach, and trainer, while the Python environment communicates with these clients through shared memory and synchronizes them at the simulator-cycle level. By preserving the original soccer-server match logic and HELIOS behavior modules, R2D-RL converts the competition-oriented workflow into a step-based learning environment. Figure~\ref{fig:architecture} summarizes this design. {Based on these components, the next subsection defines the interaction semantics exposed to MARL algorithms.}

\subsection{RL Interface and Environment Semantics}
This subsection defines the environment semantics exposed by R2D-RL to MARL algorithms, including play-mode handling, centralized state, local observations, action spaces, action masks, termination, rewards, opponent settings, and scenario initialization.

\paragraph{{Relation to standard MARL APIs.}}
{R2D-RL exposes simultaneous-action MARL semantics through a PyMARL-style interface. Gymnasium or PettingZoo integration currently requires an adapter. The current interface accepts a joint NumPy action array, returns a shared team reward, shared termination indicator, and information object, and exposes local observations, centralized state, and legal-action masks separately through \texttt{get\_obs()}, \texttt{get\_state()}, and \texttt{get\_avail\_actions()}. Its closest standard abstraction is the PettingZoo Parallel API because all active players act simultaneously \cite{terry2021pettingzoo}. An adapter would map fixed array rows to agent identifiers, convert arrays to per-agent dictionaries, include observations in \texttt{reset()} and \texttt{step()} returns, define per-agent rewards, separate termination from truncation, declare observation and action spaces, and expose legal-action masks through observation or information dictionaries.}

\paragraph{Play-mode handling.}
RCSS2D represents the current match state using play modes, including \(\mathrm{play\_on}\), kickoff, free kick, kick in, corner kick, and goal kick modes. R2D-RL preserves this native play-mode mechanism. Because \(\mathrm{play\_on}\) corresponds to the continuous open-play phase in which regular movement, passing, shooting, and defensive decisions mainly occur, R2D-RL exposes policy decision frames only during \(\mathrm{play\_on}\).

In restart modes such as kickoff, free kick, kick in, corner kick, and goal kick, the restart procedure is handled by the RCSS2D referee logic and {the built-in HELIOS decision logic}. {One counted environment step corresponds to one joint policy decision at a synchronized \(\mathrm{play\_on}\) cycle. During uninterrupted \(\mathrm{play\_on}\), the next observation normally comes from the following server cycle. If a goal or restart intervenes, the non-\(\mathrm{play\_on}\) cycles are handled internally without requesting additional policy actions or advancing the episode-step counter. After \(\mathrm{play\_on}\) resumes, the pending \texttt{step()} call returns the reward associated with the intervening event together with the first synchronized post-restart observation. Thus, one counted environment step may span multiple raw simulator cycles when a restart occurs.}

\paragraph{{Centralized state.}} The centralized state is supplied by the modified coach and is represented from the left-team perspective by default. {It is transmitted through a separate coach-side shared-memory channel and exposed through \texttt{get\_state()} for centralized-training methods.} {Decentralized actors receive only their player-side local observations at execution time.} In the current implementation, the coach state is a 136-dimensional vector, with dimension \(4 + 22 \times 6 = 136\). The first four entries encode the ball \(x/y\)-position and \(x/y\)-velocity. The remaining entries correspond to 22 player slots, covering both teams. Each player slot contains six scalar features: \(x/y\)-position, \(x/y\)-velocity, body direction, and team identifier.

\paragraph{{Local observations.}} {Each player client generates its local observation from its own HELIOS WorldModel. At each cycle, the client combines view-limited RCSS2D visual information with body-sensor information and action history to update the WorldModel estimates of itself, the ball, and other players. In the reported experiments, player-side full-state messages and Gaussian visual noise are disabled, while measurement quantization and distance-dependent attribute omissions in the standard visual protocol are retained. R2D-RL converts the updated WorldModel estimates into policy observations without adding observation noise or applying a separate denoising stage.} {For the 11-vs-11 training runs reported in Table~\ref{tab:full_field_benchmark_results}, both \texttt{quantize\_step} and \texttt{quantize\_step\_l} are set to \(10^{-6}\). The server uses synchronous and automatic modes with a 100-ms simulation cycle, one player type, relaxed stamina settings, zero ball, kick, and player randomness, no actuator noise, and no wind.} Each player observation is a flat 97-dimensional vector, with dimension \(6 + 4 + 11 \times 4 + 10 \times 4 + 3 = 97\). {The first six entries consist of the player's own \(x/y\)-position, \(x/y\)-velocity, a constant placeholder fixed to \(1.0\) by the current Python preprocessing, and the kickable flag.} {The corresponding C++ shared-memory field contains stamina.} {Current Python preprocessing replaces this entry with \(1.0\), so the policies used in the reported experiments do not observe varying stamina values.} The next four entries encode the ball \(x/y\)-position and \(x/y\)-velocity. The following entries contain up to 11 opponent slots and up to 10 teammate slots, each with four scalar features: \(x/y\)-position and \(x/y\)-velocity. {The player slots follow the native tracked-container order in the HELIOS WorldModel; the exporter does not additionally sort them by uniform number or distance.} The final three entries are scalar indicators for game mode, side, and goalie status. {Unused residual slots in the teammate and opponent blocks are zero-padded, so the observation dimension remains fixed across full-field and small-sided settings.} {For each teammate or opponent, R2D-RL writes \((x,y,v_x,v_y)\) only when the corresponding position is marked as valid in the player's HELIOS WorldModel; otherwise, the slot is zero-filled.} {Coordinates and velocities are normalized on the Python side.}

\paragraph{Actions.} The final execution layer in RCSS2D is formed by lower-level body commands. Under ordinary \(\mathrm{play\_on}\) control, the relevant body-command types include turn, dash, kick, catch, and tackle. {On top of this shared body-command execution layer, R2D-RL provides two action interfaces: Hybrid and Base. At each environment step \(t\), every controlled player selects an action, and the client sends one corresponding body command to the RCSS2D server. The two interfaces operate at the same decision frequency but differ in how the body command and its continuous arguments are determined within each environment step.}

{Building on the native RCSS2D body-command space, the Hybrid action space directly exposes a lower-level body-command interface and its parameters to the policy.} It exposes turn, dash, kick, and catch; tackle is excluded because its unstable execution outcome can severely interfere with stable possession and dribbling control during training. {Specifically, the policy outputs}
{
\begin{equation}
\label{eq:hybrid_action_output}
a_t^{\mathrm{H}}=(k_t,u_t),
\end{equation}
}
{where \(k_t\) is a discrete body-command type and \(u_t\) contains the corresponding normalized continuous arguments. This output directly specifies the body command and its arguments sent by the client at environment step \(t\). For example, the policy jointly selects kick and its associated power and angle. At the implementation level, the indexed Hybrid action space contains six IDs. Four IDs correspond to policy-selectable parameterized commands, while HELIOS Fallback and empty remain in the indexed space as internal execution IDs with fixed zero masks and therefore cannot be selected by the policy. Table~\ref{tab:hybrid_action_parameterization} summarizes all six indexed IDs and identifies the two internal entries.}

{To incorporate soccer domain knowledge, reduce the burden of directly exploring lower-level commands and continuous arguments, and support a broader range of discrete-action RL and MARL algorithms, we construct the Base action space using behavior modules from HELIOS-base.} The Base action space is a HELIOS-based high-level semantic interface. {Under Base, the policy outputs only a discrete soccer-semantic label,}
{
\begin{equation}
\label{eq:base_action_output}
a_t^{\mathrm{B}}=b_t.
\end{equation}
}
{Given the local world model \(\hat{w}_t\) at the current environment step, the corresponding HELIOS-base behavior module generates the body command and continuous arguments to be sent by the client within the same step:}
{
\begin{equation}
\label{eq:base_behavior_mapping}
(k_t,u_t)=H(b_t,\hat{w}_t).
\end{equation}
}
Tackle is retained in Base because {HELIOS-base} constrains its executability and execution, unlike directly parameterized Hybrid commands. {The original movement behaviors in HELIOS-base primarily determine target positions from team formations and are not directly suitable as directional control primitives for an RL policy. We therefore introduce four directional Move actions and four directional Dribble actions.} {The direction-labeled Move and Dribble actions are R2D-RL-defined directional primitives expressed as field-coordinate offsets under the left-team field convention. Their directions are defined in field coordinates and remain independent of the agent's body orientation.} Move actions denote short off-ball movements with a step size of \(3.0\) along the field \(x\)- or \(y\)-axis from the player's current position, while Dribble actions use the same directional offsets when the player is kickable and execute ball-carrying dribble behavior. For pass actions, the policy selects only the pass type, while the corresponding {HELIOS-base} pass behavior selects the receiver and target point internally using distance-based feasibility checks from the current world model. {The indexed Base interface likewise reserves HELIOS Fallback and empty as internal execution IDs that are unavailable to the policy.} The current Base actions are summarized in Table~\ref{tab:base_action_set}. In the reported experiments, the \texttt{advance} behavior is disabled because it can immediately move the ball over a long distance and introduce large discontinuous changes in possession and field position.

{Thus, at the same decision frequency, Hybrid allows the policy to directly control selected lower-level body commands and their continuous arguments, whereas Base provides the policy with predefined high-level soccer-semantic actions and uses HELIOS-base to translate them into lower-level body commands and continuous arguments based on the current local world model. The former is intended for direct discrete--continuous low-level control, while the latter constrains the policy's decision space through soccer semantics and HELIOS-base behavior priors. The two interfaces are complementary and non-nested, reflecting distinct control abstractions.}

\begin{table}[t]
\caption{Hybrid action parameterization in R2D-RL.}
\label{tab:hybrid_action_parameterization}
\centering
\footnotesize
\begin{tabularx}{\linewidth}{@{}lcX@{}}
\toprule
Action & ID & Parameterization \\
\midrule
turn & 0 & angle \(\theta \in [-180^\circ, 180^\circ]\) \\
dash & 1 & power \(p \in [1, 100]\). {Uses the default forward direction; negative power and an independent direction parameter are not exposed.} \\
kick & 2 & power \(p \in [0, 100]\), angle \(\theta \in [-180^\circ, 180^\circ]\) \\
catch & 3 & angle \(\theta \in [-90^\circ, 90^\circ]\) \\
HELIOS Fallback & 4 & no continuous parameter \\
empty & 5 & no continuous parameter \\
\bottomrule
\end{tabularx}
\end{table}

\begin{table}[t]
\caption{High-level discrete action set in R2D-RL.}
\label{tab:base_action_set}
\centering
\footnotesize
\begin{tabularx}{\linewidth}{@{}XXX@{}}
\toprule
\multicolumn{3}{c}{Actions} \\
\midrule
Tackle & Intercept & Shoot \\
Advance & Direct Pass & Lead Pass \\
Through Pass & Hold Ball & Catch \\
Dribble Up & Dribble Down & Dribble Left \\
Dribble Right & Move Up & Move Down \\
Move Left & Move Right & HELIOS Fallback \\
Empty & & \\
\bottomrule
\end{tabularx}
\end{table}

\paragraph{Action masks.}
{For the Base action space, the action mask inherits two mechanisms from the HELIOS-base behavior logic: action-executability checks and the shoot-priority rule.} {The executability checks mask behaviors that are inapplicable in the current state; for example, passing and dribbling require kickability, while catching requires the corresponding goalkeeper conditions.} {Under the shoot-priority rule, Shoot becomes the only selectable Base action whenever the HELIOS shooting predicate is satisfied.} {The Base-mask ablation therefore evaluates the combined effect of these two HELIOS-derived mechanisms rather than pure invalid-action removal.} For the Hybrid action space, command-level masks restrict low-level commands using their simulator-side executability conditions: kick commands are enabled only for kickable players, and catch commands are enabled only for goalies satisfying the catch-executability conditions. {In both interfaces, HELIOS Fallback and empty are fixed-masked internal entries unavailable to the policy.} Detailed Base and Hybrid action-mask conditions are summarized in Appendix~\ref{app:action_masks}, Tables~\ref{tab:base_action_mask_details} and~\ref{tab:hybrid_action_mask_details}.

\paragraph{Termination.}
R2D-RL implements configurable episode-termination conditions.
\textbf{Goal event:} the controlled team scores or concedes a goal.
\textbf{Out-of-play event:} the ball goes out of play.
\textbf{Possession loss:} the defending side is assigned ball ownership by the environment's nearest-player possession rule. This requires a defender to be within the kickable-distance threshold of the ball, with no controlled-team player equally close or closer to the ball.
\textbf{Timeout:} the episode reaches the configured horizon. {Only synchronized \(\mathrm{play\_on}\) Python environment steps advance the timeout counter; raw RCSS2D cycles during restart modes are excluded.}

\paragraph{{Scoring reward.}}
{R2D-RL uses a team-level scoring reward shared by all controlled players.} {The reward is \(+1\) when the controlled team scores, \(-1\) when it concedes a goal, and \(0\) otherwise.} {Possession-loss, out-of-play, and timeout terminations add no additional reward or penalty.}

\paragraph{{MaxEPV reward.}}
{R2D-RL optionally augments the scoring reward with a team-shared MaxEPV reward.} {Building on the EPV framework introduced by Fern\'{a}ndez et al.~\cite{fernandez2019epv}, the MaxEPV reward provides denser training feedback to alleviate the sparsity of goal-only reward signals.} {R2D-RL adopts the publicly available, position-only EPV grid \(V\) from LaurieOnTracking~\cite{laurie_on_tracking_epv} as a fixed external positional prior and maps it to the RCSS2D pitch.} {Let \(s_t=(x_t^{\mathrm{ball}},y_t^{\mathrm{ball}})\) denote the continuous ball position, and let \(c(s_t)=(i_t,j_t)\in\mathcal{C}\) denote the corresponding grid cell, where \(\mathcal{C}=\{1,\ldots,32\}\times\{1,\ldots,50\}\), \(i_t\) indexes the \(y\)-axis rows, and \(j_t\) indexes the \(x\)-axis columns.} {The positional value used by the environment is obtained by the lookup \(V(c(s_t))=V_{i_t,j_t}\).} {The source-grid values lie in \([0,0.5714]\) and remain fixed during training.} {The R2D-RL environment operationalizes this transferred prior as the MaxEPV reward using the kickability gate, {episode-level running maximum, and goal-completion rule} defined below.} {Appendix~\ref{app:epv_validation} uses HELIOS-base mirror matches to show empirically that the spatial ordering of the source EPV grid is largely preserved in RCSS2D.} {At episode reset, the lookup value at the synchronized reset position establishes the initial {episode-level} running maximum without producing a reward:}
{
\begin{equation}
\label{eq:maxepv_initialization}
b=V\!\left(c(s_{\mathrm{reset}})\right),
\qquad M_0^{\mathrm{ep}}=b.
\end{equation}
}
{Each episode contains one MaxEPV accumulation interval. In the front-goal scenarios, the maximum is reinitialized at the next scenario reset after termination by a goal, out-of-play event, possession loss, or timeout. In the 11-vs-11 benchmark, the maximum is retained throughout the 3000-step episode and is reinitialized only at the next episode reset; goals and other native match restarts within that episode do not reinitialize it. All reported training and evaluation experiments use this reset convention.}
{At environment step \(t\), let \(g_t=\mathbb{I}[\exists i\in\mathcal{I}_{\mathrm{ctrl}}:\mathrm{kickable}_{i,t}=1]\) indicate whether any controlled player is kickable.} {The gated MaxEPV reward and {episode-level} running maximum are}
{
\begin{equation}
\label{eq:maxepv_update}
\begin{aligned}
r_t^{\mathrm{EPV}}
&=g_t\alpha\max\left(0,V\!\left(c(s_t)\right)-M_{t-1}^{\mathrm{ep}}\right),\\
M_t^{\mathrm{ep}}
&=\begin{cases}
\max\left(M_{t-1}^{\mathrm{ep}},V\!\left(c(s_t)\right)\right), & g_t=1,\\
M_{t-1}^{\mathrm{ep}}, & g_t=0.
\end{cases}
\end{aligned}
\end{equation}
}
{When \(g_t=0\), the MaxEPV reward is zero and {\(M_t^{\mathrm{ep}}\)} remains unchanged.} {When \(g_t=1\), only an EPV increase above the current {episode-level} running maximum produces a reward; moving to a lower-value location is not penalized, and returning to a previously reached value does not produce repeated reward.} {When the controlled team scores, the environment adds a MaxEPV completion reward of \(\alpha(V_{\max}-M_t^{\mathrm{ep}})\) after the gated update and then raises \(M_t^{\mathrm{ep}}\) to \(V_{\max}=0.5714\); an opponent goal produces no completion reward. Neither team's goal resets the running maximum within a nonterminating episode. After the first controlled-team goal, further MaxEPV progress and completion rewards are zero until the next episode reset; the scoring reward remains active.} {The front-goal scenarios use \(\alpha=2.0\), while the reported 11-vs-11 configuration uses \(\alpha=1.0\).}

{Using the episode reset as the accumulation boundary gives MaxEPV a bounded auxiliary reward budget for each task attempt. For the proportional-increment reward used here, the undiscounted MaxEPV return, including goal completion, telescopes to \(\alpha(M_T^{\mathrm{ep}}-b)\leq\alpha(V_{\max}-b)\). Retaining the maximum across possession changes and match restarts prevents repeated rewards for revisiting previously reached values; in particular, conceding a goal does not replenish this budget. For a fixed reset baseline \(b\), an episode containing a controlled-team goal receives total undiscounted MaxEPV reward \(\alpha(V_{\max}-b)\), regardless of the intermediate path. The consequences of this reset choice for feedback density and policy learning are discussed in Section~\ref{sec:limitations}.}

\paragraph{Built-in AI.}
{Opponent and non-controlled teammate behaviors are handled through the same internal action interface as controlled players.} {The legacy internal label ``HELIOS Fallback'' invokes the complete HELIOS-base decision pipeline distributed with R2D-RL, rather than a simplified default routine.} {In the 30M-step 11-vs-11 experiments, all 11 opposing players execute this pipeline.} {In the scenario experiments, every active defender likewise executes the complete HELIOS-base controller.} {Inactive players are frozen and do not participate in decision making, and Empty Goal is a 1-vs-0 scenario with no opponent.} {The learning policy selects actions only for the controlled players.}

\paragraph{Scenario reset.}
R2D-RL supports both built-in benchmark start states and user-defined scenario initialization. The built-in start states are used for the reported front-goal and full-field experiments, while custom scenarios can specify the initial ball state and player states, including position, body direction, and velocity.

{The following subsection explains how R2D-RL realizes these interface semantics across the independently executing RCSS2D server and clients.}

\subsection{{Step-Synchronous Execution Protocol}}

{A shared-memory sequence-counter protocol coordinates the Python environment with the separate RCSS2D clients and maps each \(\mathrm{step}(\mathrm{actions})\) call to a synchronized \(\mathrm{play\_on}\) decision point.} {Trainer-side reset requests use a separate request--acknowledgment mechanism.} Without changing the soccer server's match logic, the protocol introduces a synchronization boundary between the modified HELIOS clients and the Python environment: controlled players publish decision points, the environment writes actions for those decision points, and the players acknowledge that the actions have been consumed.

{The player-side protocol uses the sequence fields summarized in Table~\ref{tab:player_sequence_fields}.} One synchronized environment step proceeds as follows.

\begin{table}[!ht]
\caption{Player-side sequence fields.}
\label{tab:player_sequence_fields}
\centering
\renewcommand{\arraystretch}{1.08}
\begin{tabular}{@{}ll@{}}
\toprule
Field & Meaning \\
\midrule
\(\mathrm{phase}\) & Current play mode. \\
\(\mathrm{cycle}\) & Player-side simulator cycle. \\
\(\mathrm{obs\_seq}\) & Observation-publication sequence. \\
\(\mathrm{act\_seq}\) & Action-assignment sequence. \\
\(\mathrm{done\_seq}\) & Action-consumption sequence. \\
\bottomrule
\end{tabular}
\end{table}

\begin{enumerate}
\item Each controlled player in the \(\mathrm{play\_on}\) phase writes its current observation and action mask to the player shared-memory buffer, then increments \(\mathrm{obs\_seq}\).
\item The environment waits until the participating players satisfy \(\mathrm{phase}=\mathrm{play\_on}\) and \(\mathrm{obs\_seq} > \mathrm{act\_seq}\).
\item The environment reads the current player observations, action masks, and coach-side centralized state, and exposes them to the learning algorithm.
\item The learning algorithm returns a joint action for the controlled players.
\item The environment writes each selected action to the corresponding player buffer and sets \(\mathrm{act\_seq}\) to the target \(\mathrm{obs\_seq}\).
\item Each player waits until \(\mathrm{act\_seq}=\mathrm{obs\_seq}\), reads the assigned action, and invokes the corresponding HELIOS behavior or low-level command.
\item After consuming the action, each player updates \(\mathrm{done\_seq}\). The environment waits until the participating players satisfy \(\mathrm{done\_seq}=\mathrm{act\_seq}\), confirming that the joint action has been consumed.
\end{enumerate}

{Together with RCSS2D's native synchronization mode, the sequence-counter barrier ties each counted environment step to a synchronized \(\mathrm{play\_on}\) decision point and aligns the coach-side centralized state with the player-side observations from that decision point.}

\paragraph{{Logical-cycle and wall-clock timing.}}
{The server retains a nominal 100-ms logical cycle and runs in synchronous mode.} {During training, R2D-RL extends the server- and client-side completion waits beyond the native per-cycle deadline, while a 300-s Python PlayOn-readiness watchdog provides a fault-recovery boundary.} {Consequently, policy inference that exceeds 100 ms can remain associated with the same logical cycle.} {This preserves the association between each policy action and its resulting API transition, as well as the observation--action--transition ordering, while relaxing wall-clock fidelity relative to competition timing.} {Across 1,000 completed 11-vs-11 decision cycles, the complete-path median, 95th-percentile, and 99th-percentile latencies were 6.029, 8.728, and 12.748 ms, respectively.} {The maximum latency was 412.392 ms because of two simulator/player ready-state recovery events.} {These measurements characterize latency under training-time synchronization.} {Competition deployment requires explicit deadlines, fallback actions, and deadline-miss reporting to establish hard real-time compliance.}

{For the 11-vs-11 experiments, the environment checks the status of the native processes and ensures that the coach state is up to date with the player observations. If synchronization fails, the affected transition is excluded from learning and the environment must be reset or restarted.}

\paragraph{{Reset synchronization.}}
The trainer client uses a \(\mathrm{req\_seq}\)/\(\mathrm{done\_seq}\) request-acknowledgment mechanism for reset and scenario initialization, bringing reset operations into the same synchronization design. At \(\mathrm{reset}()\), the Python environment selects a built-in or custom start state and sends it to the trainer through shared memory. {To avoid the per-episode cost of restarting the soccer server, the trainer resets the running server state by directly moving the ball and players to the requested states.} After repositioning, the environment waits for {a configured number of warm-up simulator cycles} so that player world models and observations are updated consistently. {In the reported 11-vs-11 experiments, this setting is one cycle for MAPPO and IPPO and three cycles for QMIX and ParaDQN.} Once the trainer acknowledgement and synchronized \(\mathrm{play\_on}\) observations are available, the environment clears episode-level bookkeeping and returns the first observation for rollout. Together, these mechanisms convert the original competition-oriented execution into repeatable step-synchronous MARL transitions.

\section{Experiments and Results}\label{sec:experiments}
The experiments evaluate R2D-RL from three complementary perspectives: parallel sampling throughput, progressive front-goal scenarios, and full-field 11-vs-11 benchmarks.

\subsection{{Algorithms}}

{For the Base action space, we evaluate Multi-Agent Proximal Policy Optimization (MAPPO), Independent Proximal Policy Optimization (IPPO), and QMIX \cite{yu2022surprising,rashid2018qmix}. MAPPO and IPPO use the same parameter-sharing recurrent PPO architecture and optimization settings. Their actors receive player-local observations augmented with controlled-player identifiers; the MAPPO critic receives the independent 136-dimensional coach-side centralized state, whereas the IPPO critic receives the same player-local input as its actor. For the Hybrid action space, we use an independent parameterized deep Q-network (ParaDQN) adapted from the single-agent parameterized-action formulation \cite{xiong2018parametrized}. ParaDQN is used as a parameter-sharing independent learner: all controlled agents share one Q network and one parameter actor, while each agent selects its own parameterized Hybrid action from its local observation. The parameter actor outputs the continuous action parameters, and the Q network selects the discrete Hybrid command conditioned on those parameters. No centralized state input, centralized critic, or value decomposition is used. Its role in this study is to validate the trainability of the Hybrid action interface {in the single-attacker scenarios} under a parameter-sharing independent learner. Detailed algorithm training settings are provided in Appendix~\ref{app:baseline_settings}, Tables~\ref{tab:common_training_settings},~\ref{tab:baseline_algorithm_architectures}, and~\ref{tab:baseline_optimization_settings}. {Hyperparameters follow the implementation-specific configurations and documented task-level adjustments; no common systematic hyperparameter search was used across methods.}}

\subsection{{Sampling and Training Throughput}}
{In practical reinforcement-learning training, parallel environments are commonly used to improve rollout-collection efficiency.} {We therefore measure how the environment-sampling throughput of R2D-RL changes with the number of parallel environments.} {One joint environment step denotes one joint action decision at the R2D-RL API level: all controlled agents submit one joint action, and the environment returns the next observations, reward, and termination signal.}

{Measurements were conducted using one socket of a dual-socket AMD EPYC 7542 node (32 physical cores and 64 logical CPUs), with all benchmark processes restricted to that socket.} {For each configuration from 2 to 20 parallel environments, one fixed-seed run measured 50,000 joint environment steps; timing began after the initial reset and included mid-run resets.} {Game and text logging were disabled.} {Actions were sampled uniformly from the legal-action mask, and Hybrid continuous arguments were sampled uniformly from \([0,1]\).} {The Base and Hybrid runs used identical environment and opponent configurations.} {Policy inference, training-data operations, and learner updates were excluded, so the measurements characterize environment and interface sampling throughput under random legal actions.} {Under the evaluated hardware configuration, these rates provide an upper-bound sampling reference for end-to-end training throughput.}

\begin{figure}[!t]
\centering
\includegraphics[width=\linewidth]{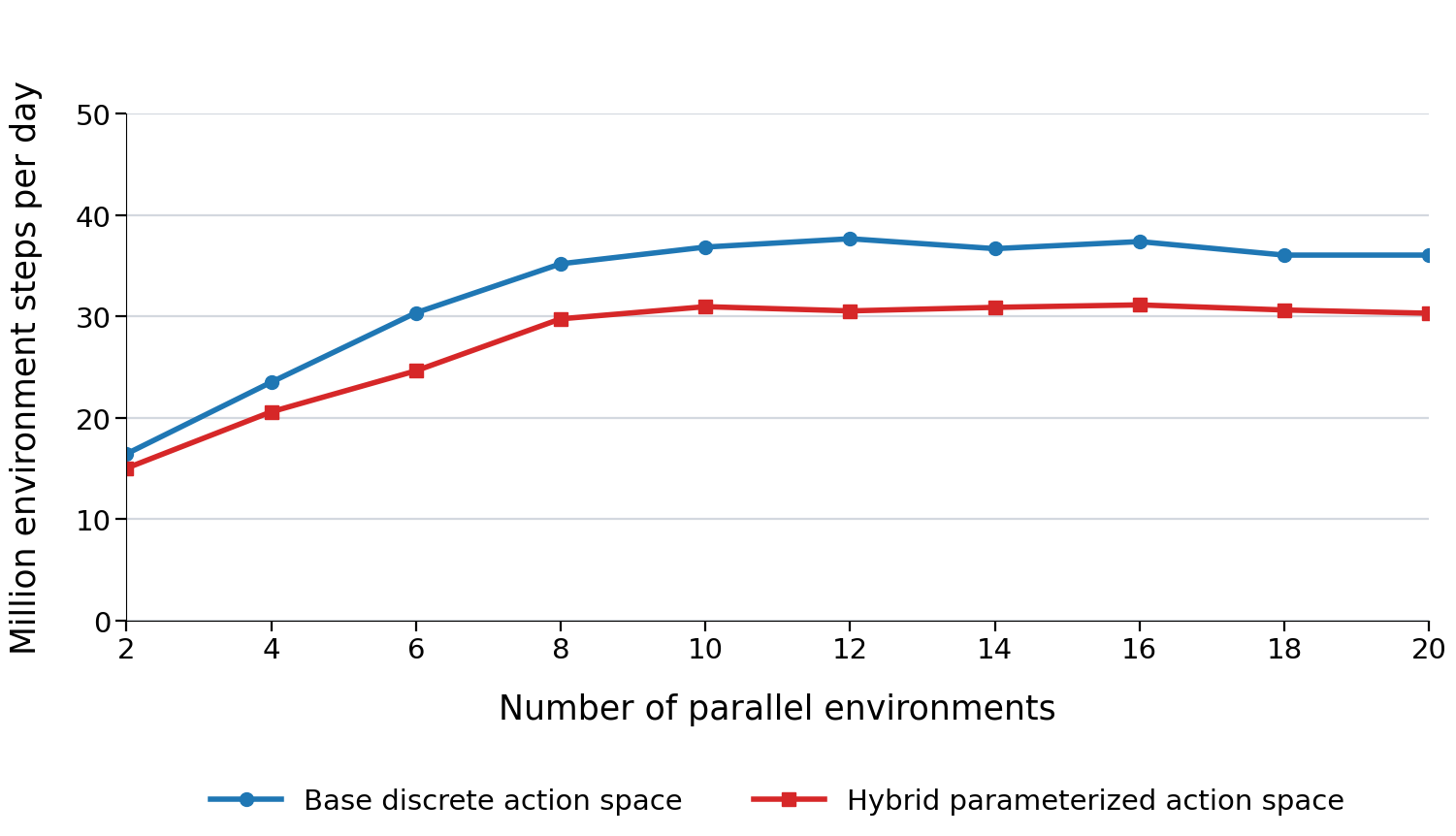}
\caption{{Parallel sampling throughput of R2D-RL in 11-vs-11 full matches under the Base and Hybrid action interfaces.}}
\label{fig:parallel_throughput}
\end{figure}

{Figure~\ref{fig:parallel_throughput} shows that aggregate throughput initially increases with parallelism and then enters a broad plateau between approximately 10 and 16 environments.} {Base records its highest observed throughput at 12 parallel environments, with 435.96 joint environment steps/s or 37.67 million steps per day.} {Hybrid records its highest observed throughput at 16 parallel environments, with 360.42 joint environment steps/s or 31.14 million steps per day.} {Several configurations within this plateau differ from the observed peaks by only a few percent, so these measurements do not identify a unique optimal parallelism level.} {Across the tested parallelism levels, Hybrid throughput is 8.58\%--18.90\% lower than the matched Base throughput.}

{Using the same single-socket allocation and 12 parallel 11-vs-11 environments, we profile MAPPO, IPPO, QMIX, and ParaDQN over three runs with seeds 0, 1, and 2.} {After warm-up, timing follows each method's configured interaction and update schedule and includes policy inference, environment interaction, buffer processing, return or target computation, learner updates, and mid-run resets; setup and the initial reset are excluded.} {Table~\ref{tab:matched_training_throughput} reports joint environment-step throughput and equivalent daily rates, together with action-space-matched uniform-random references that include legal-action sampling and environment interaction.}

\begin{table}[!t]
\caption{{Matched 11-vs-11 interaction and training throughput with 12 parallel environments. Values are means over three independent profiling runs.}}
\label{tab:matched_training_throughput}
\centering
\scriptsize
{
\setlength{\tabcolsep}{2pt}
\begin{tabular}{@{}lccc@{}}
\toprule
Profile & Action & Joint steps/s & Equiv. M/day \\
\midrule
Uniform random & Base & 427.48 & 36.94 \\
MAPPO & Base & 154.66 & 13.36 \\
IPPO & Base & 155.52 & 13.44 \\
QMIX & Base & 330.09 & 28.52 \\
\midrule
Uniform random & Hybrid & 359.49 & 31.06 \\
ParaDQN & Hybrid & 132.83 & 11.48 \\
\bottomrule
\end{tabular}
}
\end{table}
\begin{figure*}[!t]
\centering
\begin{tabular}{cc}
\includegraphics[width=0.20\textwidth]{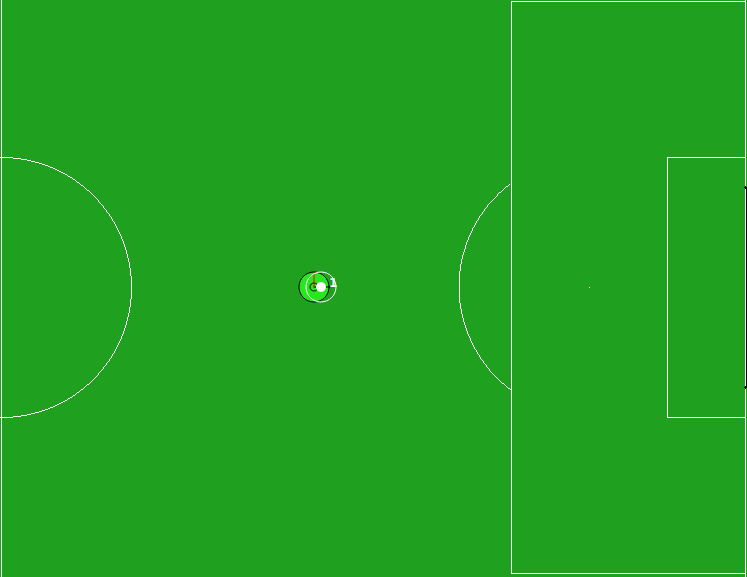} &
\includegraphics[width=0.20\textwidth]{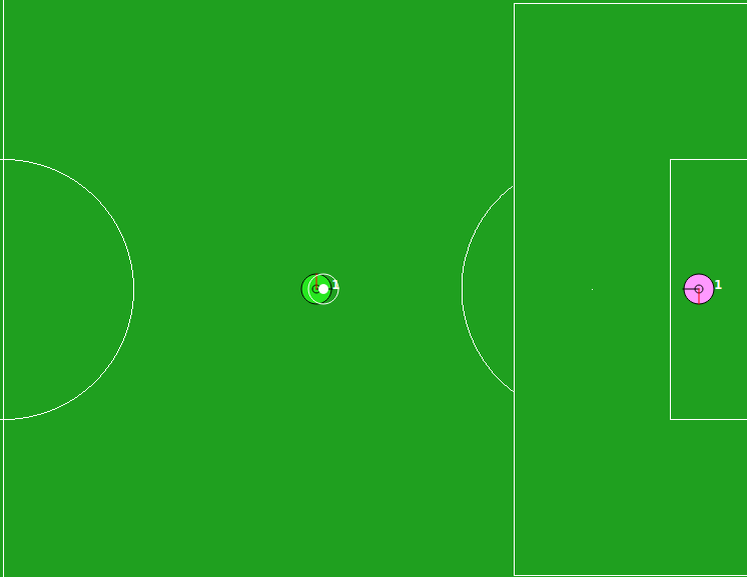} \\
(a) Empty Goal & (b) Blocked Shot \\[0.3em]
\end{tabular}

\begin{tabular}{ccc}
\includegraphics[width=0.20\textwidth]{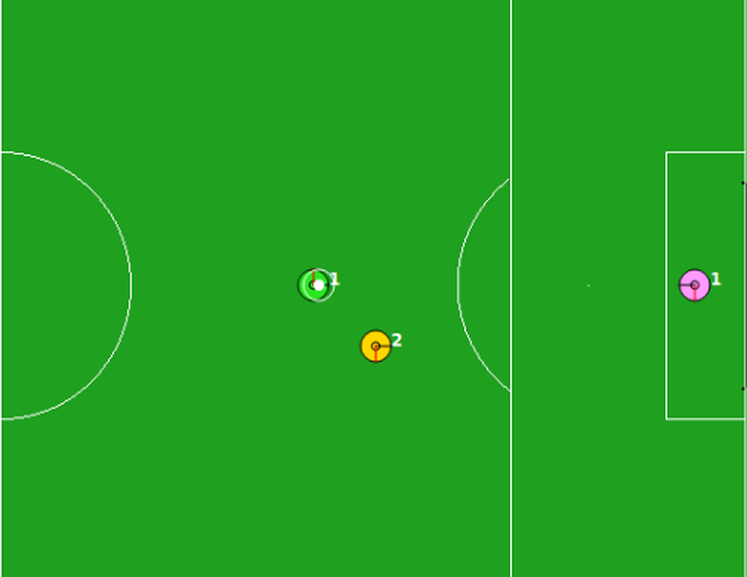} &
\includegraphics[width=0.20\textwidth]{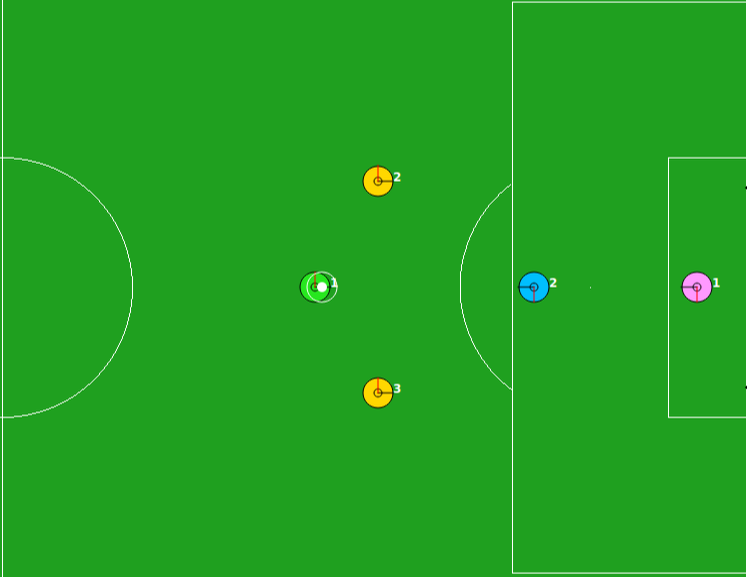} &
\includegraphics[width=0.20\textwidth]{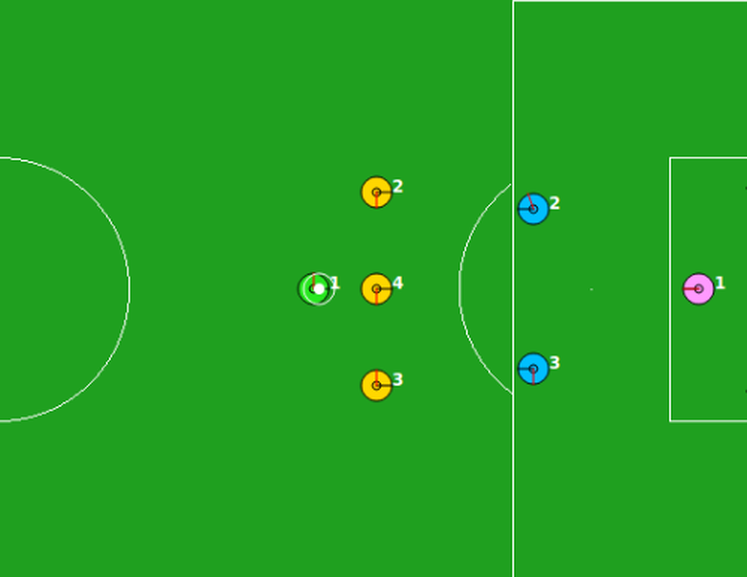} \\
(c) Support Option & (d) Passing Lane & (e) Compact Defense
\end{tabular}

\caption{Initial frames of the front-goal scenarios. Yellow and green players denote attacking players, with the green player carrying the ball. Blue players denote defending players, and the pink player denotes the goalkeeper.}
\label{fig:front_goal_catalog}
\end{figure*}
\subsection{Front-Goal Scenarios}

{Before evaluating full-field match performance, we introduce five controlled front-goal scenarios that share an attacking direction toward the right goal and form a progression from single-attacker finishing to multi-attacker coordination under increasing defensive pressure.} {This progression allows us to examine how coordination demands affect performance and how the MaxEPV reward and action-mask intervention influence learning across algorithms and scenarios.}

As shown by the initial frames in Fig.~\ref{fig:front_goal_catalog}(a)--(e), the five scenarios are Empty Goal (a), Blocked Shot (b), Support Option (c), Passing Lane (d), and Compact Defense (e). Table~\ref{tab:front_goal_catalog} summarizes their main design focus. The attacking players are controlled by the learning policy, while {all active defending players and goalkeepers execute the complete HELIOS-base decision pipeline; inactive players are frozen, and Empty Goal contains no opponent.} {Across all five scenarios, we evaluate MaxEPV+mask, mask only, MaxEPV only, and no EPV/mask; all settings use the team-level scoring reward, and goal rates are summarized over three random seeds.} {All reported QMIX front-goal runs use eight parallel environments and the same QMIX learning hyperparameters; each seed's final checkpoint is evaluated over 50 episodes.}

{Whenever a scenario is reset for a new episode, its MaxEPV running maximum is reinitialized from the synchronized reset ball position.}

\begin{table}[!htbp]
\caption{Design purposes of front-goal scenarios.}
\label{tab:front_goal_catalog}
\centering
\footnotesize
\begin{tabular}{@{}ll@{}}
\toprule
Scenario & Main focus \\
\midrule
Empty Goal & Ball progression and finishing \\
Blocked Shot & Pressured finishing \\
Support Option & Teammate support \\
Passing Lane & Off-ball support \\
Compact Defense & Coordinated attack under compact defense \\
\bottomrule
\end{tabular}
\end{table}

\subsection{Single-Attacker Front-Goal Scenarios}

We first evaluate the basic trainability of R2D-RL in two single-attacker front-goal scenarios: Empty Goal and Blocked Shot. Empty Goal is a 1-vs-0 finishing setting, while Blocked Shot adds direct defensive pressure as a 1-vs-1 setting. These scenarios are designed to verify that the environment interface, action execution, reset mechanism, and evaluation pipeline can support stable reinforcement-learning training. {They also test whether both the Base discrete action space and the Hybrid parameterized action space are trainable in simple finishing and pressured finishing settings, and whether the MaxEPV reward and action masks affect learning behavior in these settings.}

Each episode starts with the attacking side in possession. Empty Goal has a maximum horizon of 200 environment steps, and Blocked Shot has a maximum horizon of 300 environment steps. The enabled episode-termination conditions are goal event, out-of-play event, possession loss, and timeout. Both scenarios are trained for 3M environment steps with three random seeds. {Although these scenarios contain only one controlled attacker, we retain MAPPO, IPPO, and QMIX for the Base action space to keep the algorithm interface, action space, training procedure, and evaluation protocol consistent with the later multi-attacker and full-field benchmark.} For the Hybrid action space, we use the ParaDQN setting described above.

\begin{figure*}[t]
\centering
\includegraphics[width=0.78\textwidth]{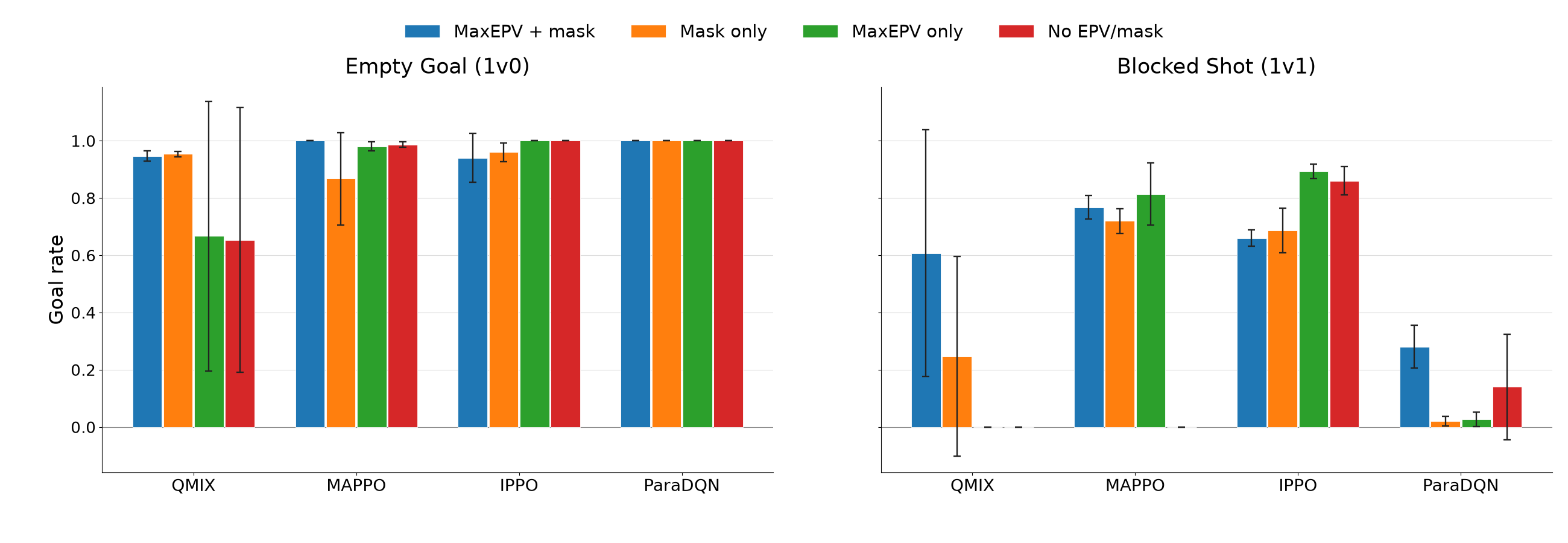}
\caption{{Single-attacker front-goal results for QMIX, MAPPO, IPPO, and ParaDQN. Bars show the goal rate over evaluation episodes, summarized as mean and population standard deviation over three random seeds. Detailed numerical values are reported in Appendix Table~\ref{tab:front_goal_detailed_results}.}}
\label{fig:single_attacker_results}
\end{figure*}

Figure~\ref{fig:single_attacker_results} and Appendix Table~\ref{tab:front_goal_detailed_results} report the single-attacker goal rates. {In Empty Goal, QMIX obtains \(0.95 \pm 0.02\) with MaxEPV+mask and \(0.95 \pm 0.01\) with mask only, whereas its two mask-free settings obtain \(0.67 \pm 0.47\) and \(0.65 \pm 0.46\). The nearly identical mask-enabled means indicate little additional win-rate benefit from MaxEPV in this simple setting, while disabling the mask substantially increases the observed variation.} {In Blocked Shot, QMIX obtains its highest mean of \(0.61 \pm 0.43\) with MaxEPV+mask, compared with \(0.25 \pm 0.35\) with mask only and zero in both mask-free settings; its MaxEPV+mask seed rates of \(0.96\), \(0.86\), and \(0.00\) show substantial seed-level variation.} {MAPPO reaches its highest mean of \(0.81 \pm {0.11}\) with MaxEPV only.} {For IPPO with MaxEPV enabled, the observed mean goal rate is \(0.89 \pm 0.02\) without the mask and \(0.66 \pm 0.03\) with the mask.} {ParaDQN obtains its highest result of \(0.28 \pm 0.07\) with both components enabled.} Appendix Fig.~\ref{fig:blocked_shot_mask_examples} provides qualitative QMIX examples for the mask-off and mask-on cases.

\subsection{Multi-Attacker Front-Goal Scenarios}

{Support Option, Passing Lane, and Compact Defense extend the front-goal evaluation to multiple controlled attackers.} The controlled side starts in possession toward the right goal, and episodes terminate on a Goal event, an Out-of-play event, Possession loss, or Timeout as defined in Section~\ref{sec:methodology}. These scenarios require teammate support, passing-lane selection, off-ball positioning, and multi-agent credit assignment. Each episode has a maximum horizon of 300 environment steps, and each method is trained for 5M environment steps with three random seeds.

\begin{figure*}[!t]
\centering
\includegraphics[width=0.95\textwidth]{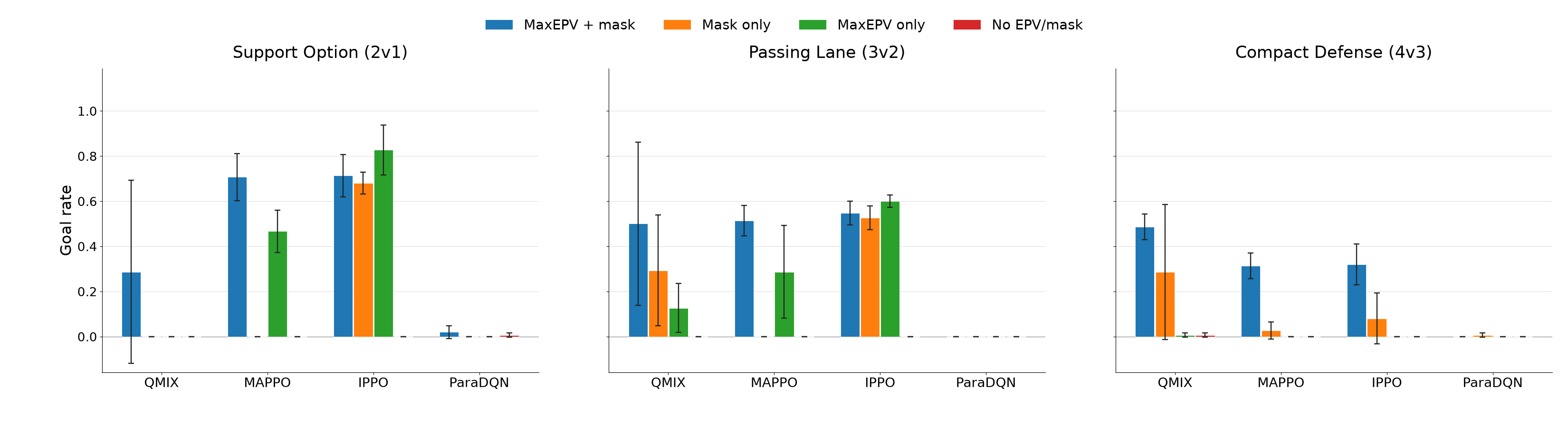}
\caption{{Multi-attacker front-goal results for QMIX, MAPPO, IPPO, and ParaDQN. Bars show the goal rate over evaluation episodes, summarized as mean and population standard deviation over three random seeds. Colors denote the four component settings: MaxEPV+mask, mask only, MaxEPV only, and no EPV/mask. Detailed numerical values are reported in Appendix Table~\ref{tab:front_goal_detailed_results}.}}
\label{fig:multi_attacker_results}
\end{figure*}

Figure~\ref{fig:multi_attacker_results} and Appendix Table~\ref{tab:front_goal_detailed_results} report the multi-attacker goal rates. {For IPPO with MaxEPV enabled, the observed mask-free and masked means are \(0.83 \pm 0.11\) and \(0.71 \pm 0.09\) in Support Option, and \(0.60 \pm 0.03\) and \(0.55 \pm 0.05\) in Passing Lane, respectively.} {Under MaxEPV+mask, MAPPO decreases from \(0.71 \pm {0.10}\) to \(0.51 \pm {0.07}\) and \(0.31 \pm {0.06}\) across the three scenarios.} {Under the same setting, QMIX obtains \(0.29 \pm 0.41\), \(0.50 \pm 0.36\), and \(0.49 \pm 0.06\), respectively. The corresponding seed-level goal rates are \((0.86,0.00,0.00)\), \((0.66,0.84,0.00)\), and \((0.42,0.48,0.56)\); thus, only Compact Defense records nonzero results in all three seeds.} {For QMIX in Compact Defense, the mask-only and MaxEPV-only results are \(0.29 \pm 0.30\) and \(0.01 \pm 0.01\), respectively.} {ParaDQN remains near zero across all three scenarios.} {Appendix Fig.~\ref{fig:front_goal_4v3_ippo_onon} provides a qualitative Compact Defense example from the IPPO MaxEPV+mask setting.}

\raggedbottom
\subsection{Full-Field Benchmark}

We further evaluate R2D-RL under full-field match dynamics in the standard 11-vs-11 RCSS2D setting. Each episode has a maximum horizon of 3000 environment steps. In the reported full-field experiments, the enabled episode-termination condition is timeout; goal events trigger native RCSS2D goal restarts but do not terminate the episode. {The MaxEPV running maximum is retained for this entire 3000-step episode and is reinitialized only at the next episode reset. Neither a goal nor an ordinary possession change refreshes it.} The learning policy controls all players on the left team, while {all 11 opposing players execute the complete HELIOS-base decision pipeline as fixed built-in AI}. To reduce fixed kickoff bias, the default kickoff side is alternated across timeout resets, while goal restarts follow the native RCSS2D rule in which the conceding side kicks off.

{The goal of this experiment is to evaluate whether R2D-RL can support end-to-end training, synchronized sampling, and standardized evaluation under complete RCSS2D match dynamics.}

{All full-field experiments use the team-level scoring reward, the MaxEPV reward, and action masks.} {The learned policies are trained for 30M environment steps using three {training} seeds. {Each final checkpoint is evaluated for 50 episodes.} {All four methods use two parallel environments per training run.} {Table~\ref{tab:full_field_benchmark_results} reports the final evaluation results of policies retrained after the synchronization fix. The corresponding training curves are shown in Fig.~\ref{fig:full_field_return_curves}.} We also evaluate non-learning Base and Hybrid references that sample uniformly from the currently legal {discrete} actions under the corresponding interface. {Hybrid continuous parameters are also sampled uniformly.} Each reference uses policy seeds 0, 1, and 2 and is evaluated for 50 episodes per seed, with the simulator seed fixed at {1793161509}. These references characterize uninformed legal-action selection and the initial exploration burden associated with each interface.} We use goal difference as the primary evaluation metric and report max\_epv\_improvement as an auxiliary metric for ball progression toward higher-value attacking regions. Figure~\ref{fig:full_field_settings} shows a left-team kickoff example for the full-field benchmark.

\begin{center}
\begin{minipage}{\columnwidth}
\centering
\includegraphics[width=0.8\columnwidth]{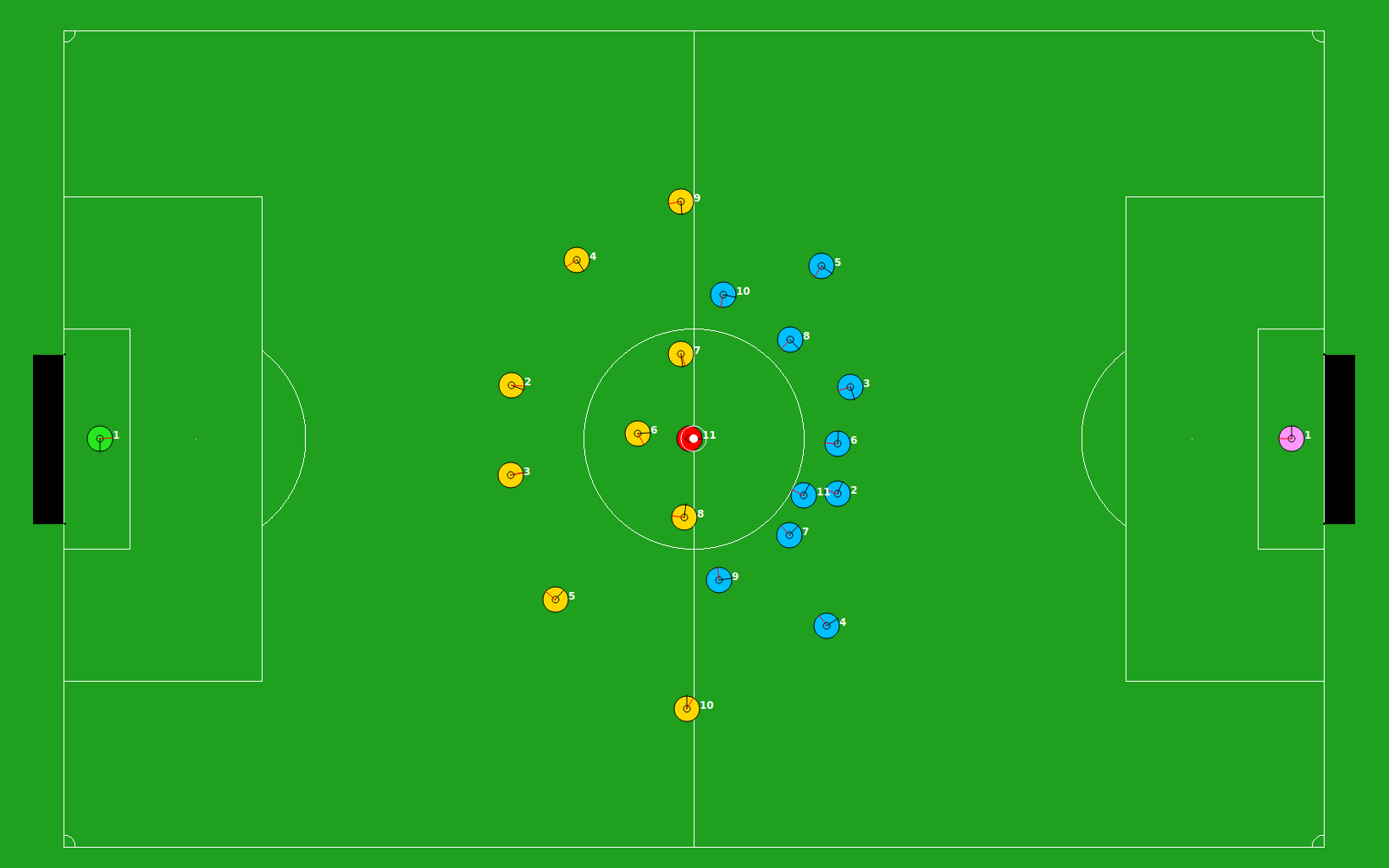}
\captionof{figure}{Left-team kickoff example for the 11-vs-11 full-field benchmark in R2D-RL.}
\label{fig:full_field_settings}
\end{minipage}
\end{center}

\begin{center}
\captionof{table}{{11-vs-11 full-field results for 30M-step learned policies and uniform-random references{, with a scripted HELIOS-base reference}. {Except for the scripted reference, values} are mean and {population} standard deviation over three {seed-level} averages, with 50 evaluation episodes per seed. {Seeds refer to training seeds for learned policies and action-sampling seeds for uniform-random references.}}}
\label{tab:full_field_benchmark_results}
\scriptsize
\begin{tabular}{@{}lcc@{}}
\toprule
Algorithm & Goal diff. & max\_epv\_improvement \\
\midrule
{Uniform random (Base)} & {\(-19.47 \pm 0.21\)} & {\(0.00194 \pm 0.00012\)} \\
MAPPO @30M & {\(-12.83 \pm 3.68\)} & {\(0.02385 \pm 0.00603\)} \\
{IPPO @30M} & {\(-12.71 \pm 2.71\)} & {\(0.03461 \pm 0.02309\)} \\
QMIX @30M & {\(-22.78 \pm 3.16\)} & {\(0.00653 \pm 0.00650\)} \\
\midrule
{Uniform random (Hybrid)} & {\(-31.18 \pm 0.03\)} & {\(0.00117 \pm 0.00001\)} \\
{ParaDQN @30M} & {\(-26.22 \pm 2.11\)} & {\(0.001095 \pm 0.000134\)} \\
\midrule
{HELIOS-base (scripted)$^{*}$} & {\(0.04 \pm 1.41\)} & {\(0.40379 \pm 0.22156\)} \\
\bottomrule
\end{tabular}
\par\smallskip
\parbox{\linewidth}{\scriptsize{$^{*}$The scripted reference reports the mean and population standard deviation across 150 individual games, not training-seed means. It comes from a separate historical evaluation protocol and provides a descriptive reference, not a strictly protocol-matched comparison.}}
\end{center}

{Goal difference is computed per episode as the number of goals scored by the controlled left team minus the number scored by the opponent.} {Except for the scripted reference, the table} reports the mean and standard deviation over the three seed-level averages. {The auxiliary metric max\_epv\_improvement is the recorded episode-level EPV maximum minus the initial EPV value; it is distinct from cumulative MaxEPV reward.} The metric is aggregated in the same way as goal difference.

{For the learned policies, we reconstruct undiscounted cumulative MaxEPV reward from logged evaluation returns as \(S_e=R_e^{\mathrm{team}}-(G_e^{+}-G_e^{-})\), where \(G_e^{+}\) and \(G_e^{-}\) are goals scored and conceded in episode \(e\). These events contribute \(+1\) and \(-1\), respectively, and MaxEPV is the only other nonzero reward component. ParaDQN's logged return is divided by 11 to recover the team scale. The aggregate reward-magnitude contribution, \(\sum_e S_e/\sum_e(G_e^{+}+G_e^{-}+S_e)\), is 0.185109\%, 0.270802\%, 0.028663\%, and 0.004175\% for MAPPO, IPPO, QMIX, and ParaDQN, respectively. This is a ratio of aggregate components across evaluation episodes, not a fraction of signed net return. The MaxEPV component is reconstructed rather than independently logged at each step.}

{The uniform policies characterize the outcome of uninformed legal-action selection and the initial exploration burden of each action interface.} {{MAPPO, IPPO, and ParaDQN reach} a less negative mean goal difference than the uniform reference for {their} own interface{, whereas QMIX does not}.} {The Base-action policies also record higher attacking-progress values than their shared reference, while ParaDQN shows {a slightly lower value than} the Hybrid reference on this metric.} {Interpretation is restricted to each method and its corresponding action-space reference.} Additional full-field results in Appendix~\ref{app:full_field_results} include the corresponding return curves in Fig.~\ref{fig:full_field_return_curves} and qualitative rollout examples in {Figs.~\ref{fig:full_field_ippo_pass_sequence} and~\ref{fig:full_field_offball_move}}.

\subsection{Discussion}

\paragraph{{Interpretation of Experimental Results.}}

The experiments show that R2D-RL supports a progressive football MARL benchmark. {Each 11-vs-11 environment comprises an independent RCSS2D server, 22 player clients, a trainer, a coach, and a Python worker. At low parallelism, waiting periods in one simulation can overlap with work in others, whereas process scheduling, memory contention, inter-process communication, and slow-environment tail latency increasingly offset this benefit as more environments share the same CPU socket. These costs help explain the observed throughput plateau. The lower Hybrid throughput is consistent with its longer action-processing path, including continuous-argument generation, shared-memory transfer, parameter validation, and command mapping. Adding policy inference, buffer processing, and learner updates further reduces throughput relative to the uniform-random references. Because the methods use different architectures and update schedules, these measurements characterize implementation cost rather than task performance.}

{Empty Goal and Blocked Shot verify the end-to-end observation, action, reward, masking, reset, and evaluation pipeline, as well as the basic trainability of the Base and Hybrid interfaces. Blocked Shot additionally introduces HELIOS-base-controlled defensive pressure and state-dependent action availability. The representative QMIX trajectories in Appendix Fig.~\ref{fig:blocked_shot_mask_examples} illustrate why masking matters: without masking, the policy may continue dribbling or moving forward after shooting becomes executable, until the ball enters the goalkeeper's catchable area.} {The shoot-priority mask removes these competing actions by making Shoot the only selectable Base action when the HELIOS shooting condition is satisfied, thereby directly guiding the transition from progression to shooting.}

{The front-goal suite then serves two complementary purposes: it provides a descriptive progression from basic finishing to coordinated multi-agent attack, and it assesses MaxEPV and action masking as the coordination demands increase. The decline in the highest observed goal rates from Support Option through Passing Lane to Compact Defense is consistent with the added requirements of passing-lane selection, off-ball support, and attack under compact defensive pressure.} {Because the Base mask inherits both action-executability checks and shoot-priority guidance from HELIOS-base, the ablation measures their joint effect.} {With MaxEPV enabled, IPPO obtains higher mask-free means in Blocked Shot, Support Option, and Passing Lane, showing that the effect of masking varies across algorithms and scenarios.} {MaxEPV provides intermediate ball-progression feedback under sparse goal success.} {However, because it is position-only and does not directly represent pass quality, off-ball support, or shooting timing, it cannot fully replace action-availability constraints.}

{Algorithmic differences provide plausible explanations for the remaining patterns. QMIX combines the Base discrete interface with monotonic value decomposition, so masking directly reduces the candidate actions considered during \(Q\)-value maximization. Its substantial seed variation suggests that off-policy value learning under sparse success remains sensitive to whether useful trajectories are discovered early; the monotonic mixing constraint may also restrict coordination whose utility depends strongly on teammates' simultaneous choices. The variation observed for the PPO methods in several settings may be related to stochastic on-policy sampling, which continues to cover alternative movement and passing choices, and to the clipped objective, which limits abrupt policy changes. Local observations may already contain sufficient task-relevant information in these small scenarios. ParaDQN faces a larger search problem because it must jointly explore lower-level command types, continuous arguments, and cross-player coordination. Under shared rewards and sparse goal success, this may prevent it from discovering effective coordinated attacks within the evaluated budget. These explanations are hypotheses supported by the observed patterns rather than isolated causal conclusions, and the results do not establish a categorical ranking among algorithms.}

{The uniform-random references reveal different starting points under uninformed action selection. Base produces soccer-semantic actions through HELIOS-base behavior modules, whereas Hybrid requires joint selection of lower-level command types and continuous arguments. Their performance gap therefore characterizes different initial exploration burdens rather than a learned algorithmic advantage.}

{{Across 150 final evaluation episodes per algorithm, MAPPO scored two goals and IPPO scored three, whereas QMIX and ParaDQN scored none. Goal difference therefore} primarily reflects goals conceded, while max\_epv\_improvement measures progress toward higher-value attacking regions. {MAPPO, IPPO, and ParaDQN achieve} a less negative goal difference than the uniform reference for {their} own interface{, whereas QMIX does not}. The Base-action policies also record higher attacking-progress values than their reference, whereas ParaDQN shows {a slightly lower value than} the Hybrid reference on this metric. {The IPPO example illustrates ball recovery, passing, and forward progression, as shown in Appendix Fig.~\ref{fig:full_field_ippo_pass_sequence}.}}

{{Limited scoring} and {the lack of} coordinated off-ball movement may be related to the current reward structure and cold-start training. Scoring and MaxEPV rewards are shared across all controlled players and provide no role-specific credit to on-ball and off-ball behavior. Without a scored goal, the scoring term supplies only concession penalties, while MaxEPV directly rewards ball progression but not the delayed contribution of distant positioning. Qualitative inspection is consistent with this credit-assignment limitation. ParaDQN players often remain nearly stationary or turn in place, whereas MAPPO and IPPO assign similar probabilities to multiple directional Move actions. Stochastic sampling during training alternates these directions and approximately preserves the formation, but greedy evaluation repeatedly amplifies small preferences and produces persistent movement in one or two directions. QMIX shows a related fixed-direction pattern under \(Q\)-value argmax execution, as illustrated in Appendix Fig.~\ref{fig:full_field_offball_move}. The evaluated policies also lack clear penalty-area-specific goalkeeper behavior.}

{With three random seeds, these quantitative patterns are descriptive and have limited statistical power.}

\paragraph{{Comparison with existing football learning environments.}} {As shown in Table~\ref{tab:football_env_comparison}, GRF and R2D-RL both support full matches, scenario-based tasks, and built-in AI, making GRF the most directly comparable environment.} {GRF uses an integrated, physics-based 3D Football Engine and derives pixel, Super Mini Map, and structured observations from the simulator state \cite{kurach2020gfootball}; this facilitates standardized comparison but abstracts away player-specific uncertainty from sensing, localization, and tracking.} {R2D-RL retains the distributed native RCSS2D architecture: each player client derives a local observation from its own HELIOS \texttt{WorldModel}, and the coach state is exposed separately for centralized training, making local uncertainty and inter-player information asymmetry part of the learning problem.}

{GRF's 19 default high-level discrete actions focus the policy on soccer decisions and provide a common interface for algorithm comparison.} {Base similarly delegates body-command parameterization to HELIOS-base, reducing low-level exploration and supporting discrete-action methods while constraining the policy to HELIOS-mediated behaviors.} {Hybrid directly exposes selected RCSS2D body commands and normalized continuous arguments, enabling parameterized-control research while increasing exploration and optimization difficulty.} {State-dependent action-availability masks reduce exploration of unavailable actions, and the modular HELIOS-base behaviors support controlled studies of rule-based priors.}

{The standard GRF scenario interface configures initial positions, player roles, and controllability for curriculum construction; R2D-RL additionally supports body orientations and velocities for precise state reconstruction and targeted diagnostics.} {These tradeoffs position GRF as an integrated algorithm and representation benchmark and R2D-RL as a platform for distributed local sensing, centralized training with decentralized execution, behavior priors, and parameterized low-level control.}

{This architecture also entails limitations. Launching and synchronizing multiple native processes adds installation, process-management, resource-management, and parallelization overhead. R2D-RL currently uses a PyMARL-style API; integration with native Gymnasium or PettingZoo conventions requires an adapter. It does not provide RGB or spatial-map learning observations or 3D rendering and currently offers a smaller built-in scenario suite and fewer reported baseline implementations than GRF. A controlled same-hardware throughput comparison remains future work. R2D-RL is positioned as a competition-oriented complement to GRF.}

\section{Limitations and Future Work}\label{sec:limitations}

{In benchmark settings that include an opponent, the current version evaluates learned policies only against the fixed, complete HELIOS-base controller and does not yet support alternative rule-based teams, learned opponents, learned-policy self-play training, or opponent pools.} {This stationary setting enables reproducible comparisons but risks opponent-specific overfitting and does not measure cross-opponent generalization. Stronger opponents would provide a stricter test and may reduce the current goal-difference and attacking-progress metrics, although this effect remains unmeasured. Future versions will support additional RCSS2D base teams and pools containing fixed teams, learned policies, and earlier checkpoints.} {Learned-policy self-play and opponent-pool training may improve adaptation and long-horizon coordination, but their non-stationarity can also increase training variance and sample complexity.}

{The reported policies are initial baselines under standardized training budgets, not estimates of the performance ceiling attainable in R2D-RL. Additional training may provide limited gains in the near-ceiling single-attacker tasks but could help the multi-attacker and 11-vs-11 policies discover passing, formation maintenance, and attack--defense coordination under sparse rewards. More environment steps do not guarantee monotonic improvement and may instead lead to plateaus or increased opponent-specific overfitting. Future work will evaluate larger budgets together with role-aware credit assignment, formation-aware auxiliary objectives, curricula, demonstrations, or pretraining; their benefits must be measured rather than assumed.}

{The Base action space relies on HELIOS-derived high-level actions and hand-designed availability masks. These priors reduce exploration but constrain learned behavior and introduce domain-specific inductive biases. Future work will compare alternative action abstractions and mask designs, including interfaces with less hand-coded structure.}

{Learning decisions are exposed only during \(\mathrm{play\_on}\); native RCSS2D rules and HELIOS logic handle restarts and set pieces. This keeps open-play transitions synchronized but excludes important strategic situations such as kickoffs, free kicks, goal kicks, and penalty kicks. Future versions will expose selected non-\(\mathrm{play\_on}\) modes while preserving synchronization.}

{The current local observations omit the observation age and confidence maintained by the HELIOS \texttt{WorldModel}, lack explicit missingness indicators for zero-filled player slots, and replace stamina with a constant value of \(1.0\). Future versions will expose normalized stamina, observation-age, confidence, and missingness features for uncertainty-aware MARL.}

{Episode-level MaxEPV accumulation limits the renewal of auxiliary feedback. Later attacks earn shaping only by improving the episode record, so a long goalless interval can receive no further MaxEPV reward despite repeated attacking attempts. After the first controlled-team goal, MaxEPV feedback is exhausted until the next reset, although the scoring reward remains active. This limitation is especially relevant to the 3000-step 11-vs-11 benchmark, where multiple attacks share one accumulation interval, whereas front-goal scenarios begin a new interval at each scenario reset. Its effect on training performance has not been isolated.}

{The equal undiscounted MaxEPV totals for episodes in which the controlled team scores, given the same reset baseline, do not establish policy invariance: discounting can favor earlier reward delivery, and episodes in which the controlled team does not score can finish at different maxima. Goal-based resets would renew feedback but link the available auxiliary budget to goal-event frequency. Possession-based resets would provide more frequent renewal but require stable ownership detection and safeguards against repeatedly refreshing the budget through possession changes. These alternatives remain untested here, and the present results do not establish that episode-level resets are preferable. Future work will compare reset boundaries under matched training conditions and quantify how they affect reward density and the contribution of MaxEPV to episode return.}

\section{Conclusion}\label{sec:conclusion}

This paper presented R2D-RL, a cycle-synchronized MARL environment that bridges the original RCSS2D simulator, HELIOS-based player clients, coach and trainer processes, and Python learning algorithms. {R2D-RL preserves the original server-client structure and exposes it through training-oriented environment semantics, including synchronized observations, centralized state, policy-level action spaces, action masks, rewards, termination conditions, and scenario reset.} This design positions R2D-RL as a full-match MARL environment built on RCSS2D while remaining compatible with the established RoboCup 2D workflow.

The benchmark results show that R2D-RL supports progressive football MARL experiments from controlled front-goal scenarios to full 11-vs-11 match dynamics. The environment can be sampled in parallel at practical scale, the single-attacker scenarios verify the basic trainability of both Base and Hybrid action spaces, and the multi-attacker and full-field settings expose increasing difficulty in teammate use, passing-lane selection, credit assignment, and long-horizon attacking progression. {The results also show that the MaxEPV reward and action masks are useful environment components for complex football tasks, while the remaining gaps in full-field behavior highlight the need for larger-scale training, richer curricula, learned-policy self-play, and less hand-coded action abstractions.} Overall, R2D-RL provides a reproducible foundation for studying learning-based decision making in RoboCup 2D under realistic full-match dynamics.

\section*{{Code and Data Availability}}

{The R2D-RL implementation is publicly available under the Apache License 2.0 at \url{https://github.com/open-starlab/R2DRL}. The exact QMIX, MAPPO, IPPO, and ParaDQN implementations, experiment configurations, execution records, and result files used in this study are archived on Zenodo at \url{https://doi.org/10.5281/zenodo.22267690}. The archive also records the required software versions and provides instructions for reproducing the reported experiments.}


\begin{IEEEbiography}[{\includegraphics[width=1in,height=1.25in, clip,keepaspectratio]{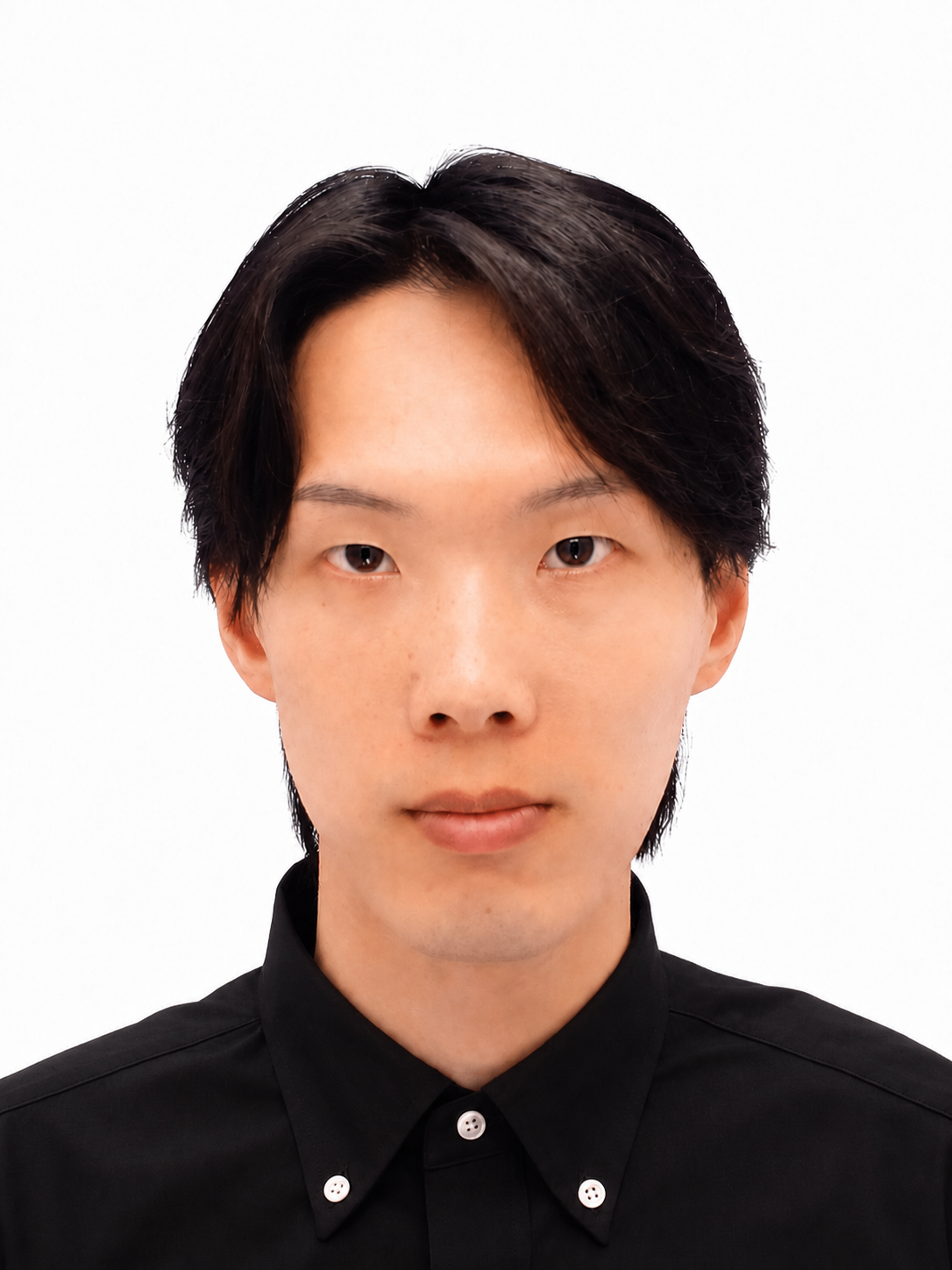}}]
{Haobin Qin}
received the B.Eng. degree from East China University of Technology and the M.Eng. degree from Beijing Institute of Technology. He is currently pursuing the Ph.D. degree at Nagoya University. His research focuses on sports intelligence, including sports simulators, virtual dataset construction, and data-driven decision-making. His research interests include reinforcement learning, computer vision, and spatio-temporal behavior modeling, with applications in sports scenarios, particularly football and basketball.
\end{IEEEbiography}

\begin{IEEEbiography}[{\includegraphics[width=1in,height=1.25in, clip,keepaspectratio]{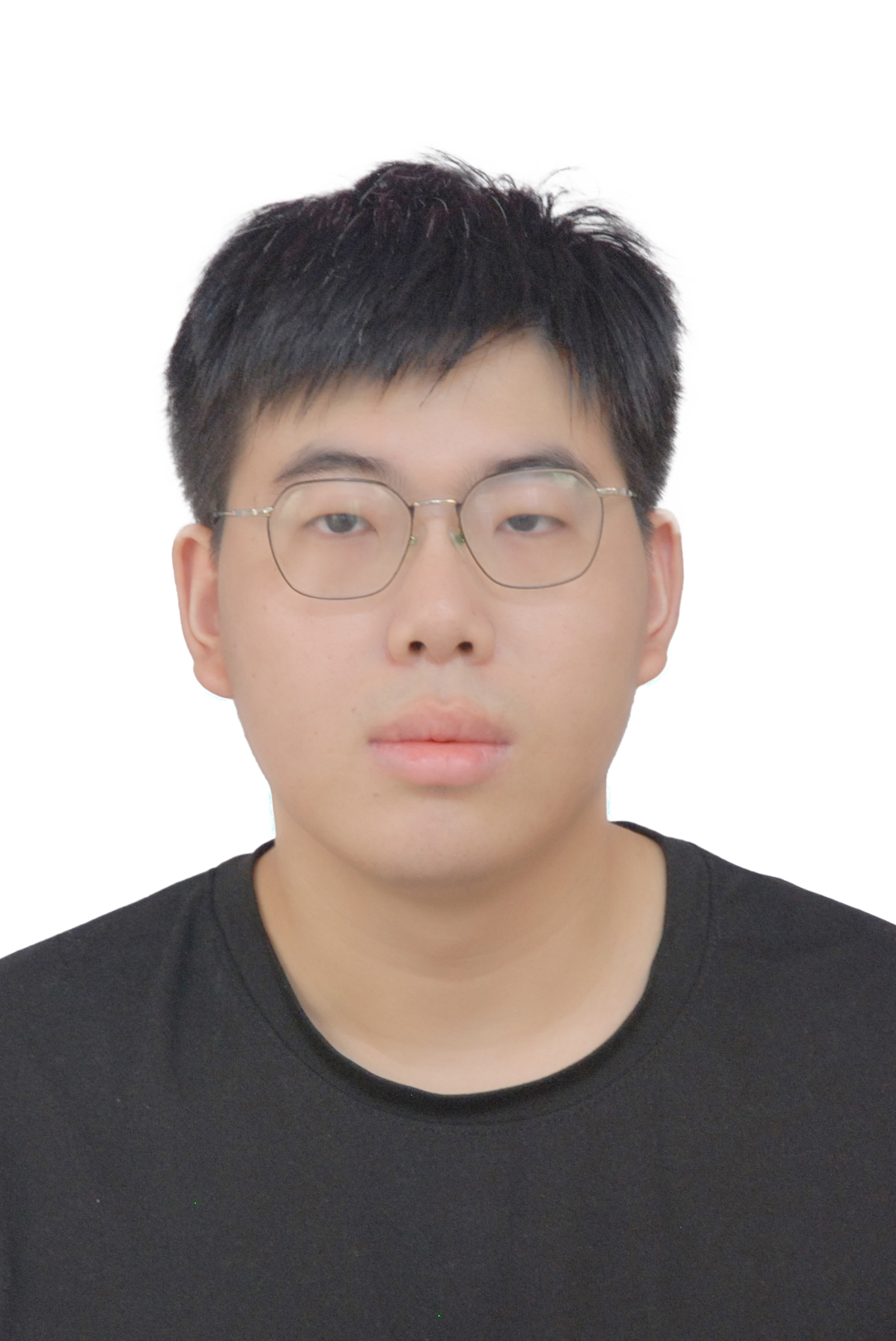}}]
{BAOFENG ZHANG}
received the B.Eng. degree in Automation from Tianjin University and the M.Eng. degree in Artificial Intelligence from the Institute of Automation, Chinese Academy of Sciences. He is currently pursuing the Ph.D. degree at Nagoya University. His research focuses on reinforcement learning and multi-agent reinforcement learning, with particular interests in game-theoretic training, learning from state-only demonstrations, and team decision-making. His current work explores sports intelligence and soccer simulation, aiming to develop data-driven and learning-based methods for multi-agent coordination in complex competitive environments.
\end{IEEEbiography}

\begin{IEEEbiography}[{\includegraphics[width=1in,height=1.25in, clip,keepaspectratio]{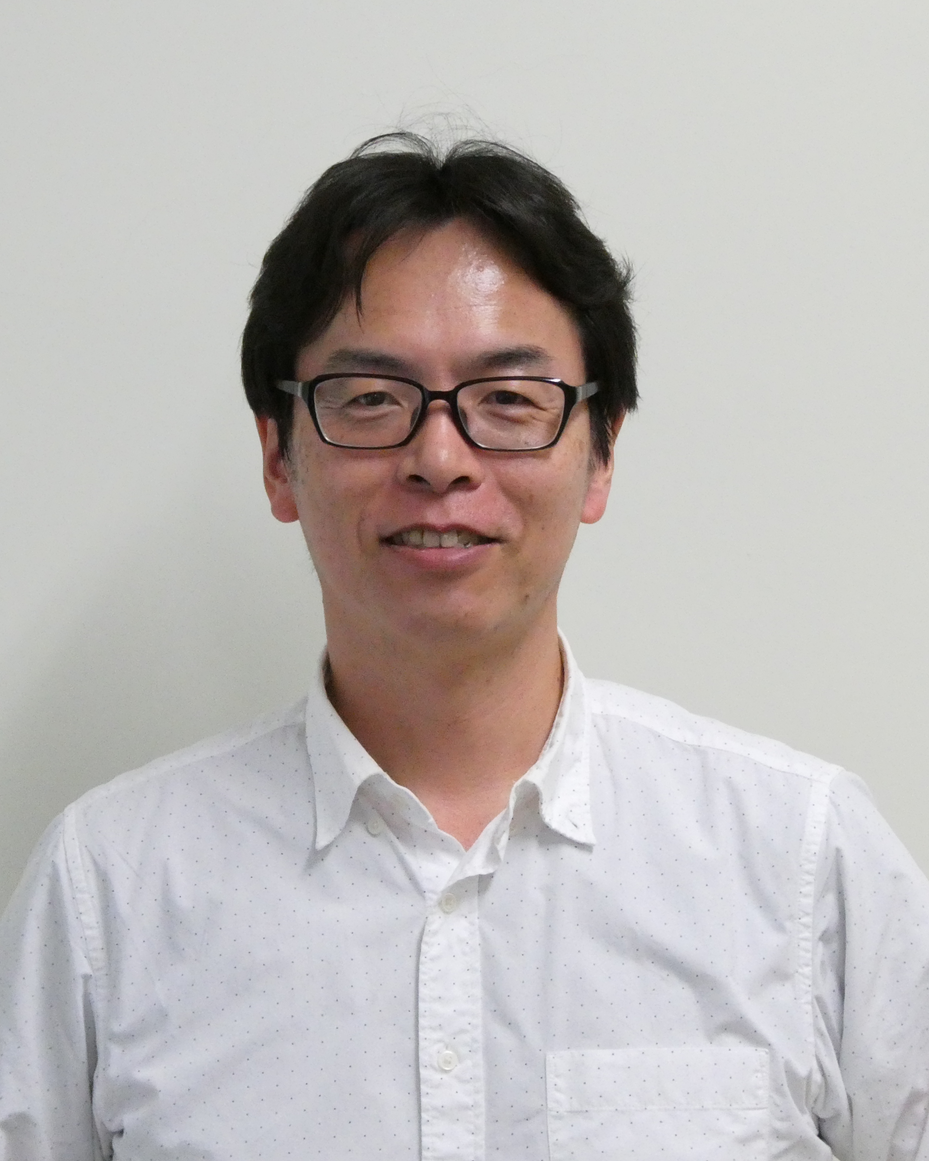}}]{Hidehisa Akiyama}
received the B.E. and M.E. degrees from Tokyo Institute of Technology, Tokyo, Japan, in 1999 and 2001, respectively, and the Ph.D. degree in engineering from Tokyo Institute of Technology in 2008.
Before joining Nagasaki University in 2026, he was an Associate Professor with the Faculty of Information Science and Engineering, Okayama University of Science, Okayama, Japan. He is currently an Associate Professor with the School of Information and Data Sciences, Nagasaki University, Nagasaki, Japan. His research interests include multi-agent systems, skill science, sports data analysis, group behavior simulation, and soccer simulation. He co-developed HELIOS Base, an open-source base team package for the RoboCup Soccer 2D Simulation, and has contributed to the RoboCup 2D soccer simulation software ecosystem.
\end{IEEEbiography}

\begin{IEEEbiography}[{\includegraphics[width=1in,height=1.25in, clip,keepaspectratio]{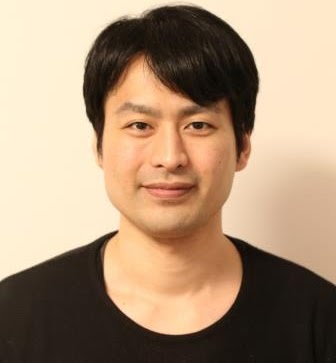}}]{Keisuke Fujii} received his B.S., M.S., and Ph.D. degrees from Kyoto University in 2009, 2011, and 2014, respectively.
After his work as a postdoctoral fellow and research scientist at Nagoya University and RIKEN Center for Advanced Intelligence Project in Japan, he joined Nagoya University.
He is currently a professor at the Information Technology Center and Graduate School of Informatics.
His research interests are interdisciplinary studies among machine learning, behavioral sciences, and sports sciences.
\end{IEEEbiography}

\clearpage
\onecolumn
\appendices
\raggedbottom
\section{Action Mask Details}\label{app:action_masks}

This appendix summarizes the action-mask logic used by R2D-RL. Action masks are generated by the modified HELIOS player client at each synchronized \(\mathrm{play\_on}\) decision frame and are passed to the Python environment through shared memory. A mask entry of \(1\) means that the corresponding policy-level action is currently selectable, while a mask entry of \(0\) means that the action is unavailable for policy selection.

For the Base action space, the mask is a 19-dimensional binary vector for each controlled player. The enabled conditions are summarized in Table~\ref{tab:base_action_mask_details}. If the shooting action is enabled, it is given priority and all other Base actions are masked out. When dynamic action masking is disabled, R2D-RL uses a static Base mask in which the normally selectable actions remain available, while advance, HELIOS Fallback, and empty remain masked out.

\begin{center}
\captionof{table}{Base action-mask conditions in R2D-RL.}
\label{tab:base_action_mask_details}
\footnotesize
\begin{tabularx}{\textwidth}{@{}c l X@{}}
\toprule
ID & Action & Enabled condition \\
\midrule
0 & tackle & HELIOS basic tackle executable; execution constrained by HELIOS. \\
1 & shoot & Kickable; not indirect free kick; not stopped; strict shoot check executable. If enabled, all other Base actions are masked out. \\
2 & intercept & Not kickable; no kickable teammate; interception logic selects this player. \\
3 & advance & Disabled in the reported experiments. \\
4 & direct pass & Kickable; direct-pass candidate; one-step kick executable. \\
5 & lead pass & Kickable; lead-pass candidate; one-step kick executable. \\
6 & through pass & Kickable; through-pass candidate; one-step kick executable. \\
7 & hold & HELIOS hold-ball behavior executable. \\
8 & catch & Goalie; not frozen; in own penalty area; valid play mode and ball position; catch ban ended; ball within catchable area minus \(0.05\). \\
9 & dribble up & Kickable. \\
10 & dribble down & Kickable. \\
11 & dribble left & Kickable. \\
12 & dribble right & Kickable. \\
13 & move up & Not kickable; \(3.0\)-m left-team field offset remains in pitch. \\
14 & move down & Not kickable; \(3.0\)-m left-team field offset remains in pitch. \\
15 & move left & Not kickable; \(3.0\)-m left-team field offset remains in pitch. \\
16 & move right & Not kickable; \(3.0\)-m left-team field offset remains in pitch. \\
17 & HELIOS Fallback & Disabled for policy selection; internal control only. \\
18 & empty & Disabled for policy selection; internal control only. \\
\bottomrule
\end{tabularx}
\end{center}

For the Hybrid action space, the mask is a 6-dimensional binary vector for each controlled player. The Hybrid mask controls which lower-level command type can be selected by the policy, while the continuous command arguments are supplied separately by the parameter heads. {Four indexed IDs are policy-selectable parameterized commands; HELIOS Fallback and empty occupy the other two indexed IDs but have fixed zero masks.} Table~\ref{tab:hybrid_action_mask_details} summarizes the current Hybrid mask.

\begin{center}
\captionof{table}{Hybrid action-mask conditions in R2D-RL.}
\label{tab:hybrid_action_mask_details}
\footnotesize
\begin{tabularx}{\textwidth}{@{}c l X@{}}
\toprule
ID & Action & Enabled condition \\
\midrule
0 & turn & The player is not in the unfrozen kickable-command branch; equivalently, the player is not kickable or is frozen. \\
1 & dash & The player is not kickable and is not frozen. \\
2 & kick & The player is kickable and is not frozen. \\
3 & catch & The same catch-executability condition as in the Base action space is satisfied. \\
4 & HELIOS Fallback & Disabled for policy selection and reserved for internal control. \\
5 & empty & Disabled for policy selection and reserved for internal control. \\
\bottomrule
\end{tabularx}
\end{center}

Thus, in ordinary non-frozen frames, the Hybrid mask separates on-ball command selection, where kick and executable catch actions can be selected, from off-ball command selection, where turn, dash, and executable catch actions can be selected.

\section{{Empirical Validation of EPV Transfer}}\label{app:epv_validation}

\paragraph{{Purpose and setup.}}
{The EPV surface used by the MaxEPV reward originates from professional-soccer tracking data, whereas RCSS2D has different player dynamics, action execution, and possession transitions.} {To examine whether the source grid retains a meaningful spatial ordering in RCSS2D, we conduct 510 independently seeded 11-vs-11 HELIOS-base mirror matches, each lasting 3000 environment steps. Both teams execute the same complete HELIOS-base decision pipeline, and no learned policy is present or updated during data collection.} {The server and all player processes are restarted for every match.}

\paragraph{{Grid mapping and possession attribution.}}
{The source EPV file contains 32 \(y\)-axis rows and 50 \(x\)-axis columns. The mapping \(c(\cdot)\) linearly assigns positions on the \(105\times68\)-m RCSS2D pitch to these 1600 cells. Both teams are expressed in a common attacking direction toward \(+x\); right-team trajectories are reflected only along the \(x\)-axis.} {A possession segment begins when a team is first identified as the owner by the true-kickable condition; simultaneous kickability is resolved using \texttt{ball\_owner\_team}. It ends when confirmed ownership changes, an explicit goal is recorded, or the match ends. Intervals without a true-kickable player do not end the segment or produce cell events, and consecutive occurrences in the same cell are deduplicated. A goal is linked to the scoring team's most recent possession ending within the preceding 200 native coach cycles.} {The resulting dataset contains 22,011 possession segments, 465 explicitly recorded goals, and 264,653 kickable events after consecutive-residence deduplication. All 465 goals are linked to a scoring-team possession trajectory.}

\begin{figure*}[!t]
\centering
\includegraphics[width=\textwidth]{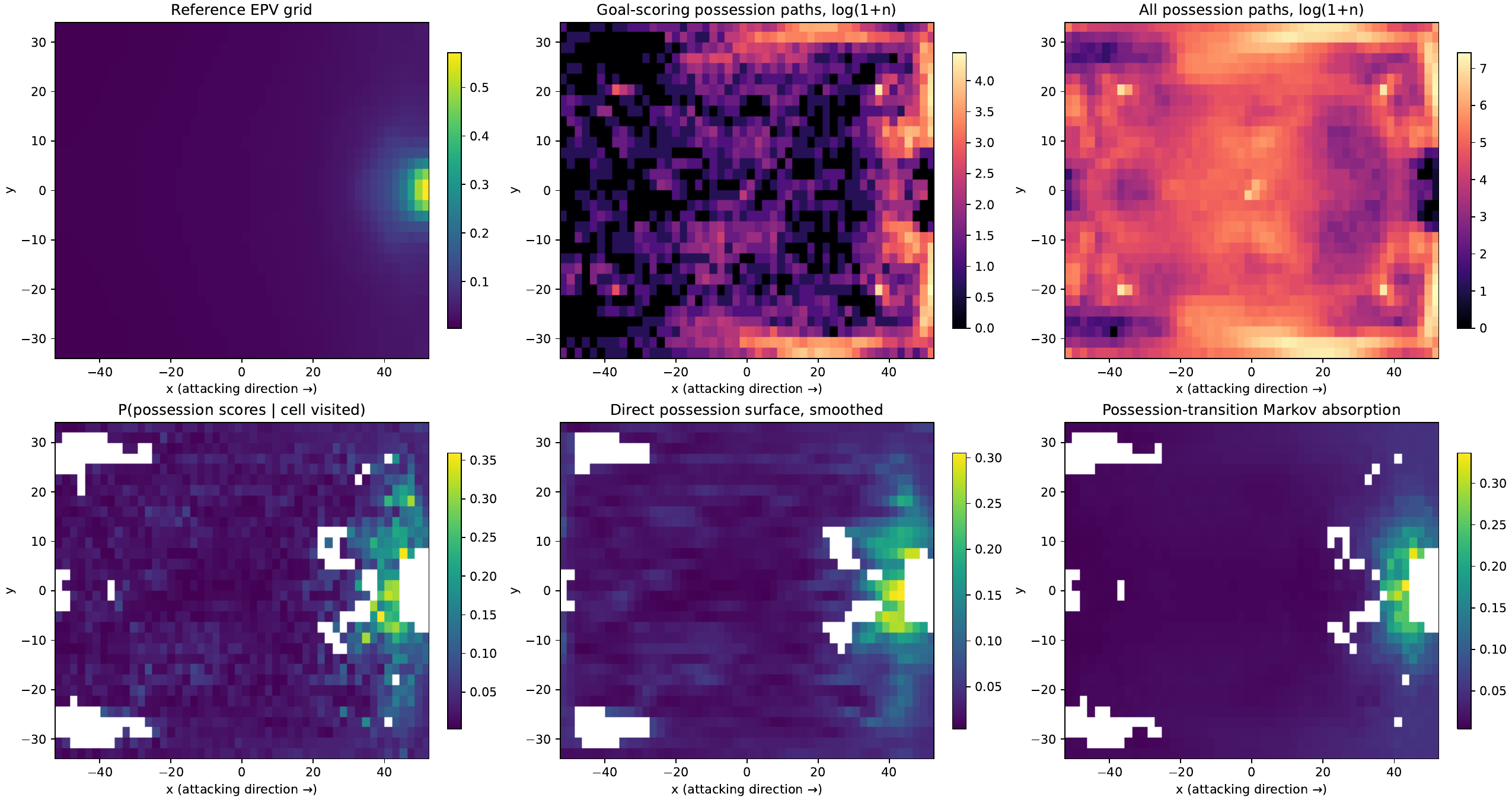}
\caption{{Reference EPV and empirical RCSS2D surfaces from 510 HELIOS-base mirror matches.} {Coordinates are expressed with the attacking direction toward positive \(x\), and each panel uses an independent color scale. White denotes cells below the support threshold of 20 in the direct, smoothed-direct, and Markov probability panels.}}
\label{fig:epv_transfer_surfaces}
\end{figure*}

\paragraph{{Empirical surfaces and comparison.}}
{To evaluate the transfer of spatial ordering, we independently estimate scoring likelihood on the same \(32\times50\) grid from the RCSS2D trajectories using two complementary methods.} {The Markov model treats the 1600 pitch cells as transient states and Goal and Failure as absorbing states: a goal-linked segment exits to Goal, while every other termination exits to the zero-valued Failure state.} {The value assigned to each cell is the estimated probability that a possession visiting that cell eventually reaches the goal state through subsequent observed transitions.} {The second method uses the terminal outcome directly and estimates, for each grid cell \(c\in\mathcal{C}\), the probability that a possession visiting the cell eventually ends in a goal.} {We apply Jeffreys smoothing to this direct estimate:}
{
\begin{equation}
\label{eq:epv_direct_validation}
\widehat{P}_{\mathrm{direct}}(\mathrm{goal}\mid c)
=\frac{N_{\mathrm{goal}}(c)+0.5}{N_{\mathrm{visit}}(c)+1}.
\end{equation}
}
{Here, \(N_{\mathrm{visit}}(c)\) is the number of possession segments that visit cell \(c\), and \(N_{\mathrm{goal}}(c)\) is the number of those segments that eventually score.} {Markov support is the total number of outgoing cell transitions and terminal exits observed for a cell; cells with support below 20 are excluded, and the retained cells are weighted by this count. The direct estimate instead uses the number of distinct possession segments visiting a cell as its support.} {Figure~\ref{fig:epv_transfer_surfaces} visualizes the source grid, trajectory densities, direct estimates, and Markov surface, while Table~\ref{tab:epv_transfer_validation} reports the weighted Pearson and Spearman correlations.} {The Gaussian-smoothed direct panel is included only to visualize broader spatial structure; the reported direct correlation uses the unsmoothed cell-level Jeffreys estimate.}

\begin{center}
\begin{minipage}{\textwidth}
\centering
{
\captionof{table}{{Empirical validation of the source EPV grid in 510 HELIOS-base mirror matches.}}
\label{tab:epv_transfer_validation}
\footnotesize
\begin{tabularx}{\textwidth}{@{}Xlrrr@{}}
\toprule
RCSS2D empirical surface & Weight & Cells & Weighted Pearson \(r\) & Weighted Spearman \(\rho\) \\
\midrule
Markov goal-absorption probability & Markov support & 1468 & 0.9306 & 0.7410 \\
Jeffreys-smoothed direct possession-to-goal estimate & Possession visits & 1454 & 0.8218 & 0.6203 \\
\bottomrule
\end{tabularx}
}
\end{minipage}
\end{center}

{Pearson \(r\) measures the linear association between cell values, while Spearman \(\rho\) measures the consistency of their rank ordering.} {Both coefficients range from \(-1\) to \(1\); values closer to \(1\) indicate stronger positive agreement, values near \(0\) indicate weak association, and negative values indicate an inverse association.} {All four coefficients in Table~\ref{tab:epv_transfer_validation} are positive: the Markov surface shows the stronger correspondence with the source grid, and the direct estimate also exhibits clear positive agreement.}

{The Markov surface remains strongly associated with the source grid after controlling for \(x\), \(x^2\), \(|y|\), and \(y^2\), with a weighted partial correlation of \(0.9369\). As a directional negative control, reflecting the source grid along the attacking \(x\)-axis produces a weighted Markov Spearman correlation of \(-0.8657\). This sign reversal demonstrates the directional dependence of the association.}

\begin{figure}[!t]
\centering
\includegraphics[width=0.5\textwidth]{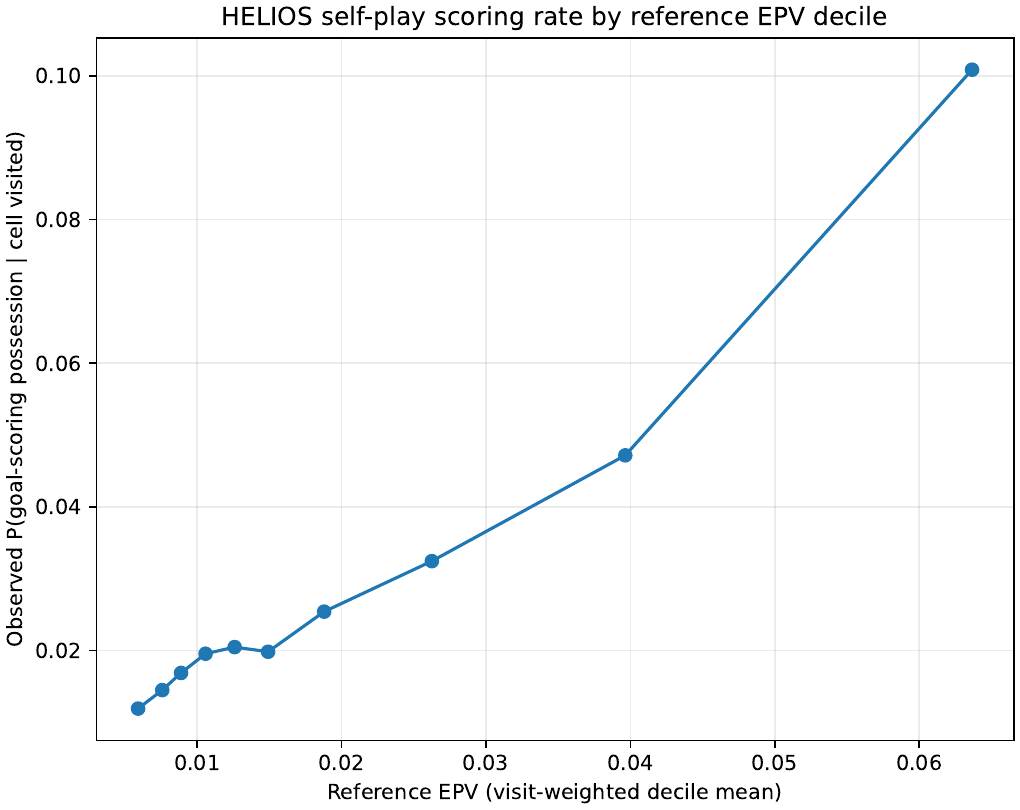}
\caption{{Observed scoring-possession rate in HELIOS-base mirror matches by reference-EPV decile.} {The plot summarizes 1454 supported cells and 234,005 possession visits and characterizes aggregate spatial ordering across deciles.}}
\label{fig:epv_decile_scoring_rate}
\end{figure}

{As a complementary aggregate view, we group the 1454 supported direct-probability cells by reference-EPV decile and compute the visit-weighted scoring-possession rate over 234,005 possession visits. Figure~\ref{fig:epv_decile_scoring_rate} shows a strong overall rise from 0.0119 in the first decile to 0.1009 in the tenth. The sixth decile is slightly below the fifth, while the overall trend remains positive across the decile range.}

\paragraph{{Interpretation.}}
{Taken together, the substantial positive correlations, the overall increase in scoring rate across reference-EPV deciles, and the directional negative control empirically validate that the spatial ordering encoded by the source EPV grid is largely preserved in RCSS2D across HELIOS-base mirror matches.} {This result supports the use of the transferred EPV grid as a fixed positional prior for the MaxEPV reward in the evaluated setting.} {The empirical scope of this validation is the spatial ranking observed in HELIOS-base mirror matches; cellwise numerical calibration and policy-independent validity beyond this setting remain open questions.}

\section{Baseline Training Settings}\label{app:baseline_settings}

This appendix summarizes the baseline training settings used in the reported experiments. The environment-side reward, termination, and action-mask semantics follow the definitions in Section~\ref{sec:methodology} and Appendix~\ref{app:action_masks}. {The number of parallel environments varies across methods and tasks; the exact settings are recorded in the released experiment configurations.}

\begin{center}
\begin{minipage}{\textwidth}
\centering
\captionof{table}{Common training and evaluation settings.}
\label{tab:common_training_settings}
\footnotesize
\begin{tabularx}{\textwidth}{@{}l X@{}}
\toprule
Setting & Value \\
\midrule
Random seeds & \(0,1,2\). \\
Episode horizon & Empty Goal: \(200\) environment steps; other front-goal scenarios: \(300\) environment steps; full-field benchmark: \(3000\) environment steps. {Only synchronized policy-decision steps issued during \(\mathrm{play\_on}\) are counted; intervening non-\(\mathrm{play\_on}\) restart cycles are excluded.} \\
Training horizon & Empty Goal and Blocked Shot: \(3\)M environment steps; other front-goal scenarios: \(5\)M environment steps; full-field benchmark: \(30\)M environment steps. \\
Evaluation & \(50\) episodes per evaluation. \\
{QMIX front-goal parallelism} & {Eight parallel environments for every component setting in all five scenarios.} \\
{Other front-goal parallelism} & {MAPPO/IPPO/ParaDQN: Empty Goal \(2/4/8\); Blocked Shot \(2/4/8\); Support Option \(2/2/2\); Passing Lane \(1/4/8\), except IPPO and ParaDQN use \(2\) when both MaxEPV and action masks are disabled; Compact Defense \(1/2/2\).} \\
Evaluation/checkpoint interval & Front-goal scenarios: every \(1\)M environment steps; full-field benchmark: every \(10\)M environment steps. \\
\bottomrule
\end{tabularx}
\end{minipage}
\end{center}

\begingroup
\begin{center}
\captionof{table}{Baseline algorithm architectures.}
\label{tab:baseline_algorithm_architectures}
\scriptsize
\setlength{\tabcolsep}{2pt}
\renewcommand{\arraystretch}{1.12}
\begin{tabularx}{\textwidth}{@{}l X X X@{}}
\toprule
Item & QMIX & {MAPPO / IPPO} & ParaDQN \\
\midrule
Action space & Base discrete action space, 19 actions. & {Base discrete action space, 19 actions.} & {Hybrid parameterized action space, 6 indexed IDs: 4 policy-selectable commands with continuous parameters and 2 fixed-masked internal IDs.} \\
Network architecture & Shared recurrent agent: FC\((o,a_{t-1},i)\)-64, ReLU, GRUCell-64, Q head; QMixer embed 32 with two-layer hypernetworks and hidden size 64. & {Separate actor and critic networks, each parameter-shared across agents; an input projection followed by two 128-unit ReLU--LayerNorm blocks and a single-layer GRU-128; categorical actor head and value critic.} & Q network: MLP \([256,256]\) over observation and full action-parameter vector; parameter actor: MLP \([256,256]\) with sigmoid-scaled continuous outputs. \\
Policy inputs & {Agent network: 97-dimensional local observation, previous-action one-hot vector, and agent-ID one-hot vector. QMixer: independent 136-dimensional coach state.} & {Actor: 97-dimensional player observation plus a one-hot controlled-player identifier (98--101 dimensions in the front-goal scenarios and 108 dimensions in 11-vs-11). MAPPO critic: independent 136-dimensional coach state. IPPO critic: the actor's local input.} & {Parameter actor: local observation plus an agent-ID one-hot vector. Q network: the same input plus the complete 12-dimensional action-parameter vector. {Two parameter slots are reserved for each of the six action IDs. Turn, Dash, and Catch each use one parameter, while Kick uses two; only the parameters of the selected command are used during execution.}} \\
Parameter sharing & Shared agent network across controlled players with centralized monotonic value mixing. & {One actor and one critic are shared across controlled players; MAPPO uses centralized value input and IPPO uses decentralized value input.} & One shared Q network and one shared parameter actor across controlled players. \\
\bottomrule
\end{tabularx}
\end{center}
\endgroup

\begingroup
\begin{center}
\captionof{table}{{Baseline optimization and evaluation settings. Task-specific settings are distinguished between the front-goal and 11-vs-11 experiments.}}
\label{tab:baseline_optimization_settings}
\scriptsize
\setlength{\tabcolsep}{2pt}
\renewcommand{\arraystretch}{1.12}
\begin{tabularx}{\textwidth}{@{}l X X X@{}}
\toprule
Item & QMIX & {MAPPO / IPPO} & ParaDQN \\
\midrule
Optimizer and learning rate & {RMSProp, \(\alpha=0.99\), \(\epsilon=10^{-5}\); learning rate: front-goal, \(5\times 10^{-4}\); 11-vs-11, \(3\times 10^{-4}\).} & Adam; actor learning rate \(3\times 10^{-4}\), critic learning rate \(5\times 10^{-4}\), \(\epsilon=10^{-5}\). & Adam; Q learning rate \(10^{-4}\), actor learning rate \(2\times 10^{-4}\). \\
Discount and returns & \(\gamma=0.99\). & \(\gamma=0.995\), GAE \(\lambda=0.98\). & \(\gamma=0.99\), 5-step return. \\
Batch and replay & {Front-goal: replay buffer 64 episodes, warm-up 32 episodes, batch size 16 episodes, and 2 learner updates per 8-episode collection batch; 11-vs-11: replay buffer 256 episodes, warm-up threshold 16 episodes, batch size 16 episodes, and 4 learner updates per {collection batch with two parallel environments after replay warm-up}.} & {On-policy rollout buffer with length equal to the environment episode limit; five PPO epochs per rollout; two recurrent minibatches per epoch; recurrent chunk length 10.} & {Replay capacity 100,000 joint environment transitions and batch size 64; one learner update per vectorized collection step in the front-goal scenarios and two in the 11-vs-11 benchmark.} \\
Target update & {Target network update every 20 episodes in the front-goal scenarios and every 100 episodes in 11-vs-11; Double Q enabled in both settings.} & Not applicable. & Double Q enabled; Polyak update with \(\tau_Q=0.003\) and \(\tau_{\mu}=0.001\); actor update every 2 Q updates. \\
Exploration & {Epsilon-greedy, annealed from \(1.0\) to \(0.05\) over 100k environment steps in the front-goal scenarios and 5M steps in 11-vs-11.} & Stochastic policy during training. & Epsilon-greedy, annealed from \(1.0\) to \(0.05\) over 1M environment steps. \\
Loss and stabilization & {Front-goal: masked MSE TD loss, full-episode BPTT, and gradient-norm clipping at 10; 11-vs-11: masked Huber TD loss with \(\delta=1\), BPTT chunk length 200, and gradient-norm clipping at 10.} & {PPO clipping coefficient 0.2; entropy coefficient 0.05; value-loss coefficient 1; clipped Huber value loss with \(\delta=10\); ValueNorm; maximum gradient norm 5; actor categorical-output-head initialization gain 0.1.} & {Huber TD loss; gradient-norm clipping at 10; actor behavior-parameter L2 coefficient 0.01; reward-biased replay sampling with \(\alpha=0.4\).} \\
Action-mask handling & Invalid actions are masked before Q maximization and action selection. & Invalid logits are set to \(-10^{10}\) before categorical sampling or evaluation. & Random exploration samples valid actions; invalid Q values are set to \(-\infty\) for Q and target-Q selection; actor loss excludes non-parameterized and fallback actions. \\
Evaluation policy & {Deterministic greedy \(Q\)-value argmax.} & {Deterministic categorical-policy argmax.} & {Deterministic greedy selection using the target actor and target Q network (\(\epsilon=0\)).} \\
\bottomrule
\end{tabularx}
\end{center}
\endgroup

\section{Detailed Front-Goal Results}
\label{app:front_goal_tables}

\begingroup
\begin{center}
\begin{minipage}{\textwidth}
\centering
\captionof{table}{Detailed front-goal results. {Values are goal rates reported as mean $\pm$ population standard deviation over three random seeds and rounded to two decimal places. Bold indicates the highest nonzero unrounded mean within each row.}}
\label{tab:front_goal_detailed_results}
\centering
\scriptsize
\setlength{\tabcolsep}{2.5pt}
\renewcommand{\arraystretch}{0.95}
\begin{tabular}{@{}llcccc@{}}
\toprule
Scenario & Model & MaxEPV+mask & Mask only & MaxEPV only & No EPV/mask \\
\midrule
\multirow{4}{*}{Empty Goal}
& {QMIX @3M}    & {$0.95 \pm 0.02$} & {$\mathbf{0.95} \pm 0.01$} & {$0.67 \pm 0.47$} & {$0.65 \pm 0.46$} \\
& {MAPPO @3M} & {$\mathbf{1.00} \pm 0.00$} & {$0.87 \pm {0.16}$} & {$0.98 \pm 0.02$} & {$0.99 \pm 0.01$} \\
& {IPPO @3M}  & {$0.94 \pm 0.08$} & {$0.96 \pm 0.03$} & {$\mathbf{1.00} \pm 0.00$} & {$\mathbf{1.00} \pm 0.00$} \\
& ParaDQN @3M & $\mathbf{1.00} \pm 0.00$ & $\mathbf{1.00} \pm 0.00$ & $\mathbf{1.00} \pm 0.00$ & $\mathbf{1.00} \pm 0.00$ \\
\midrule
\multirow{4}{*}{Blocked Shot}
& {QMIX @3M}    & {$\mathbf{0.61} \pm 0.43$} & {$0.25 \pm 0.35$} & {$0.00 \pm 0.00$} & {$0.00 \pm 0.00$} \\
& {MAPPO @3M} & {$0.77 \pm {0.04}$} & {$0.72 \pm {0.04}$} & {$\mathbf{0.81} \pm {0.11}$} & {$0.00 \pm 0.00$} \\
& {IPPO @3M}  & {$0.66 \pm 0.03$} & {$0.69 \pm 0.08$} & {$\mathbf{0.89} \pm 0.02$} & {$0.86 \pm 0.05$} \\
& ParaDQN @3M & $\mathbf{0.28} \pm 0.07$ & $0.02 \pm 0.02$ & $0.03 \pm 0.02$ & $0.14 \pm 0.18$ \\
\midrule
\multirow{4}{*}{Support Option}
& {QMIX @5M}    & {$\mathbf{0.29} \pm 0.41$} & {$0.00 \pm 0.00$} & {$0.00 \pm 0.00$} & {$0.00 \pm 0.00$} \\
& {MAPPO @5M} & {$\mathbf{0.71} \pm {0.10}$} & {$0.00 \pm 0.00$} & {$0.47 \pm {0.09}$} & {$0.00 \pm 0.00$} \\
& {IPPO @5M}  & {$0.71 \pm 0.09$} & {$0.68 \pm 0.05$} & {$\mathbf{0.83} \pm 0.11$} & {$0.00 \pm 0.00$} \\
& ParaDQN @5M & $\mathbf{0.02} \pm 0.03$ & $0.00 \pm 0.00$ & $0.00 \pm 0.00$ & $0.01 \pm 0.01$ \\
\midrule
\multirow{4}{*}{Passing Lane}
& {QMIX @5M}    & {$\mathbf{0.50} \pm 0.36$} & {$0.29 \pm 0.25$} & {$0.13 \pm 0.11$} & {$0.00 \pm 0.00$} \\
& {MAPPO @5M} & {$\mathbf{0.51} \pm {0.07}$} & {$0.00 \pm 0.00$} & {$0.29 \pm {0.21}$} & {$0.00 \pm 0.00$} \\
& {IPPO @5M}  & {$0.55 \pm 0.05$} & {$0.53 \pm 0.05$} & {$\mathbf{0.60} \pm 0.03$} & {$0.00 \pm 0.00$} \\
& ParaDQN @5M & $0.00 \pm 0.00$ & $0.00 \pm 0.00$ & $0.00 \pm 0.00$ & $0.00 \pm 0.00$ \\
\midrule
\multirow{4}{*}{Compact Defense}
& {QMIX @5M}    & {$\mathbf{0.49} \pm 0.06$} & {$0.29 \pm 0.30$} & {$0.01 \pm 0.01$} & {$0.01 \pm 0.01$} \\
& {MAPPO @5M} & {$\mathbf{0.31} \pm {0.06}$} & {$0.03 \pm {0.04}$} & {$0.00 \pm 0.00$} & {$0.00 \pm 0.00$} \\
& {IPPO @5M}  & {$\mathbf{0.32} \pm 0.09$} & {$0.08 \pm 0.11$} & {$0.00 \pm 0.00$} & {$0.00 \pm 0.00$} \\
& ParaDQN @5M & $0.00 \pm 0.00$ & $\mathbf{0.01} \pm 0.01$ & $0.00 \pm 0.00$ & $0.00 \pm 0.00$ \\
\bottomrule
\end{tabular}
\end{minipage}
\end{center}
\endgroup

\begin{center}
\begin{minipage}{0.68\textwidth}
\centering
{\scriptsize
\resizebox{0.7\textwidth}{!}{%
\begin{tabular}{cc}
\includegraphics[width=0.47\linewidth]{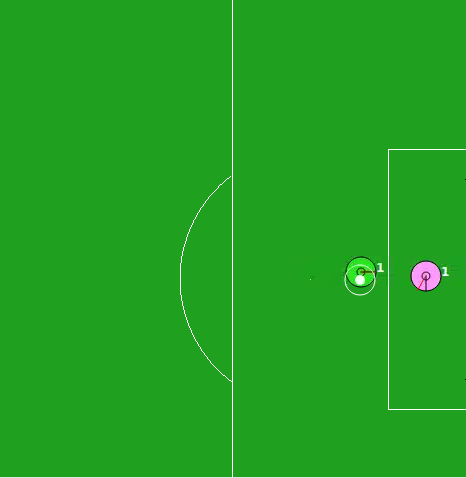} &
\includegraphics[width=0.47\linewidth]{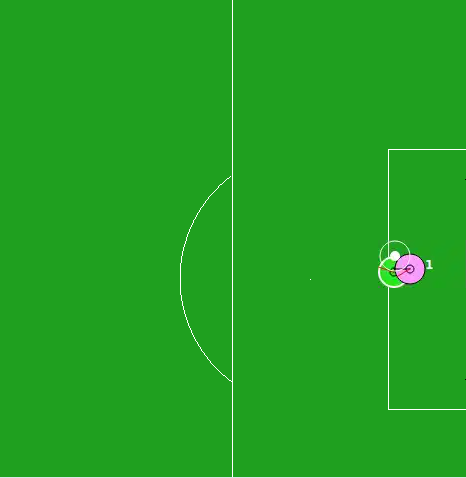} \\
(a) Mask off: goalkeeper catch & (b) Mask off: goalkeeper catch \\[0.4em]
\includegraphics[width=0.47\linewidth]{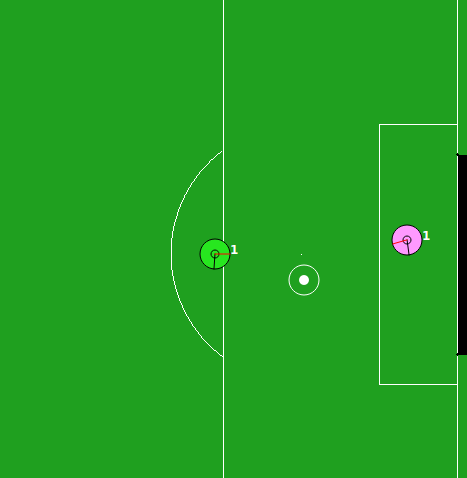} &
\includegraphics[width=0.47\linewidth]{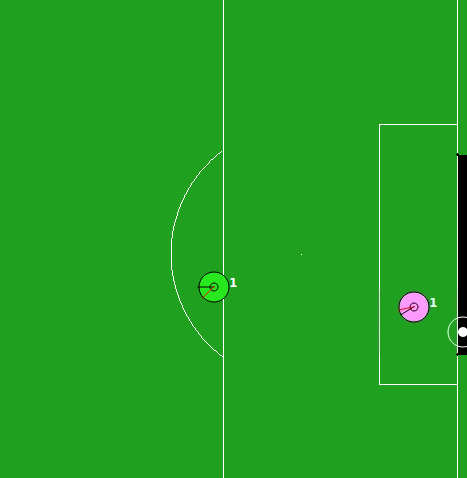} \\
(c) Mask on: shot selected & (d) Mask on: shot selected
\end{tabular}
}
}
\captionof{figure}{Qualitative QMIX examples in the Blocked Shot scenario. Green denotes the controlled attacker, pink denotes the goalkeeper, and the white circle indicates the ball. Without action masks, the policy may continue selecting dribbling or forward-movement actions until the ball enters the goalkeeper's catchable area; with the shoot-priority mask, competing actions are removed and shooting is selected when executable.}
\label{fig:blocked_shot_mask_examples}
\end{minipage}
\end{center}

\begin{center}
\begin{minipage}{0.74\textwidth}
\centering
\includegraphics[width=\linewidth]{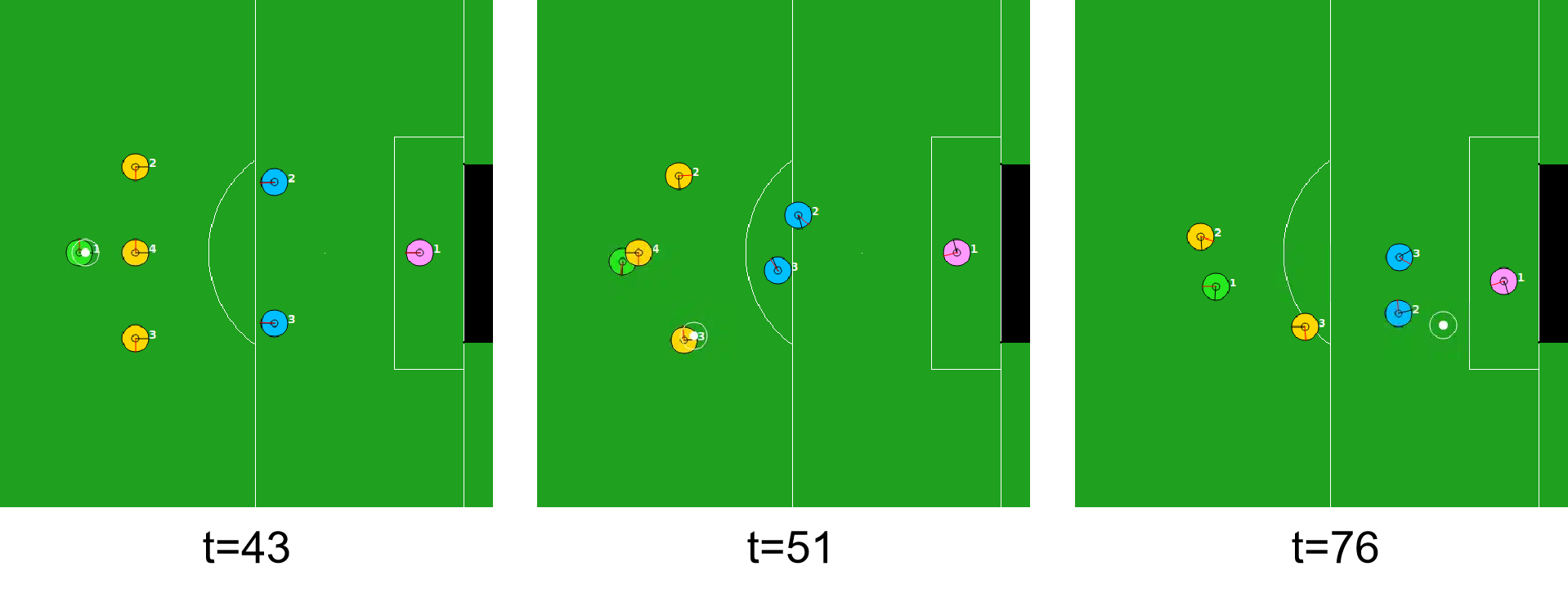}
\captionof{figure}{{Qualitative IPPO rollout in Compact Defense. Yellow and green players denote IPPO-controlled attackers, blue players denote HELIOS-base-controlled defenders, the pink player denotes the goalkeeper, and the white circle indicates the ball. The sequence is from the MaxEPV+mask setting.}}
\label{fig:front_goal_4v3_ippo_onon}
\end{minipage}
\end{center}

\section{Additional Full-Field Results}
\label{app:full_field_results}

\begin{center}
\begin{minipage}{\textwidth}
\centering
\centering
\includegraphics[width=0.7\textwidth]{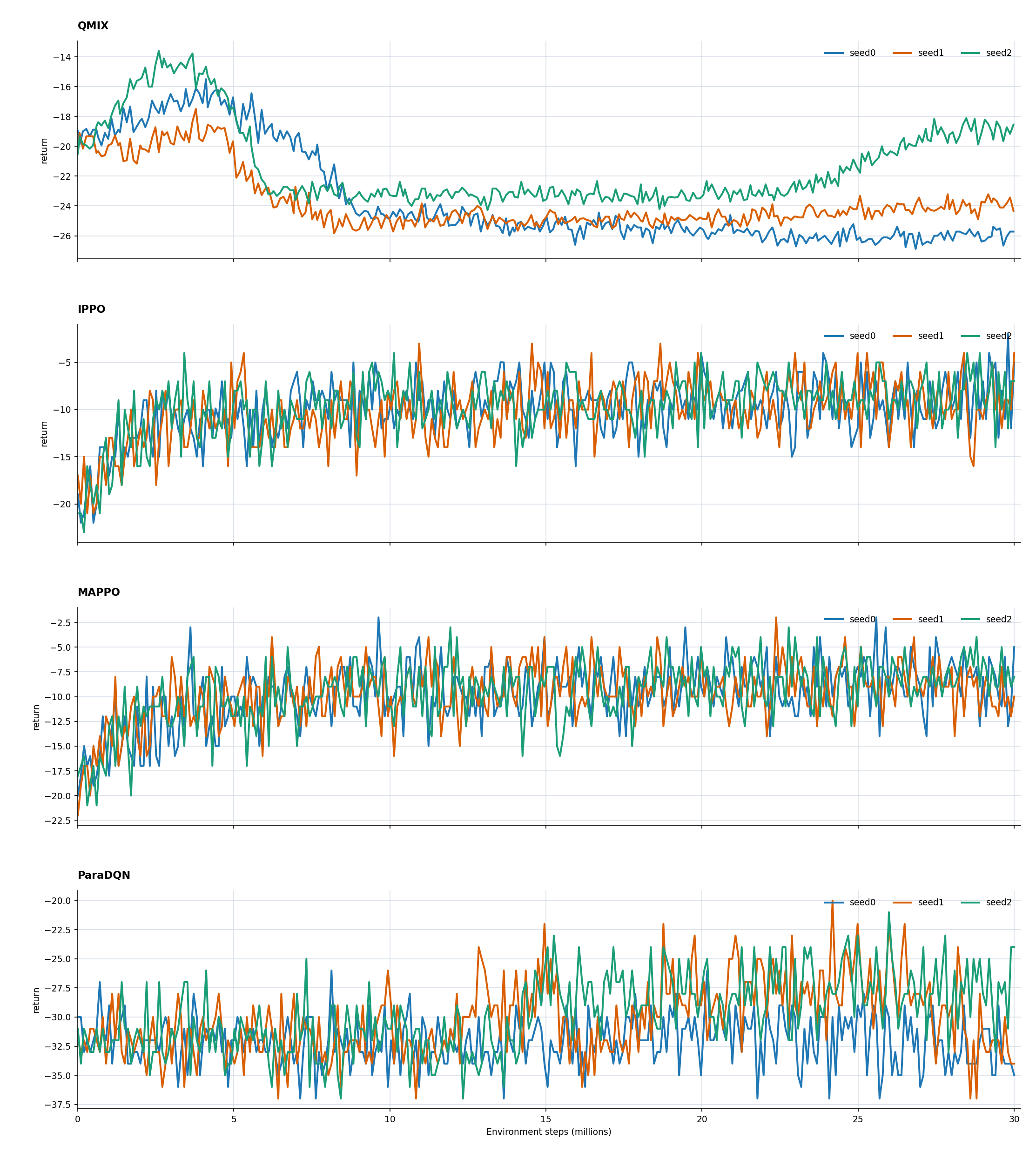}
\captionof{figure}{{11-vs-11 full-field average episode return during training over 30M environment steps. Each panel shows one algorithm, and each curve corresponds to one policy seed. {Each curve shows 300 recorded values selected at evenly spaced indices, without smoothing. ParaDQN returns are divided by 11; QMIX reports window means, whereas PPO and ParaDQN report completed-episode returns.} The panels provide within-algorithm training diagnostics under each method's own configuration.}}
\label{fig:full_field_return_curves}
\end{minipage}
\end{center}

\begin{center}
\centering
\includegraphics[width=0.62\textwidth]{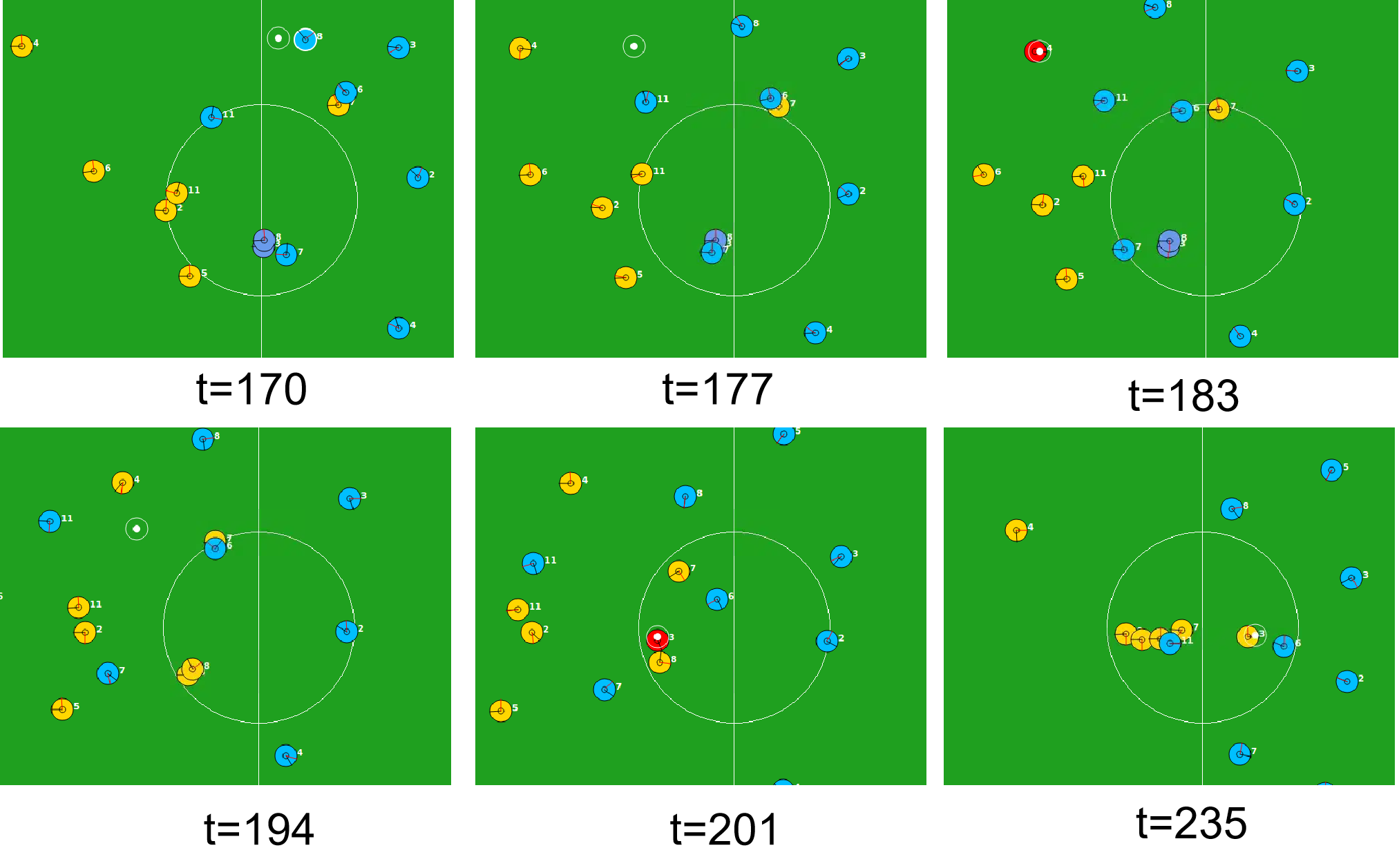}
\captionof{figure}{{Qualitative IPPO rollout example in the 11-vs-11 full-field benchmark. Yellow players are IPPO-controlled left-team agents, blue players are HELIOS-base-controlled opponents, and the white circle indicates the ball. The selected frames illustrate ball recovery, a pass, and subsequent ball progression toward the opponent half during one evaluation episode.} {This example is from an earlier 11-vs-11 evaluation.}}
\label{fig:full_field_ippo_pass_sequence}
\end{center}

\begin{center}
\centering
\includegraphics[width=0.78\textwidth]{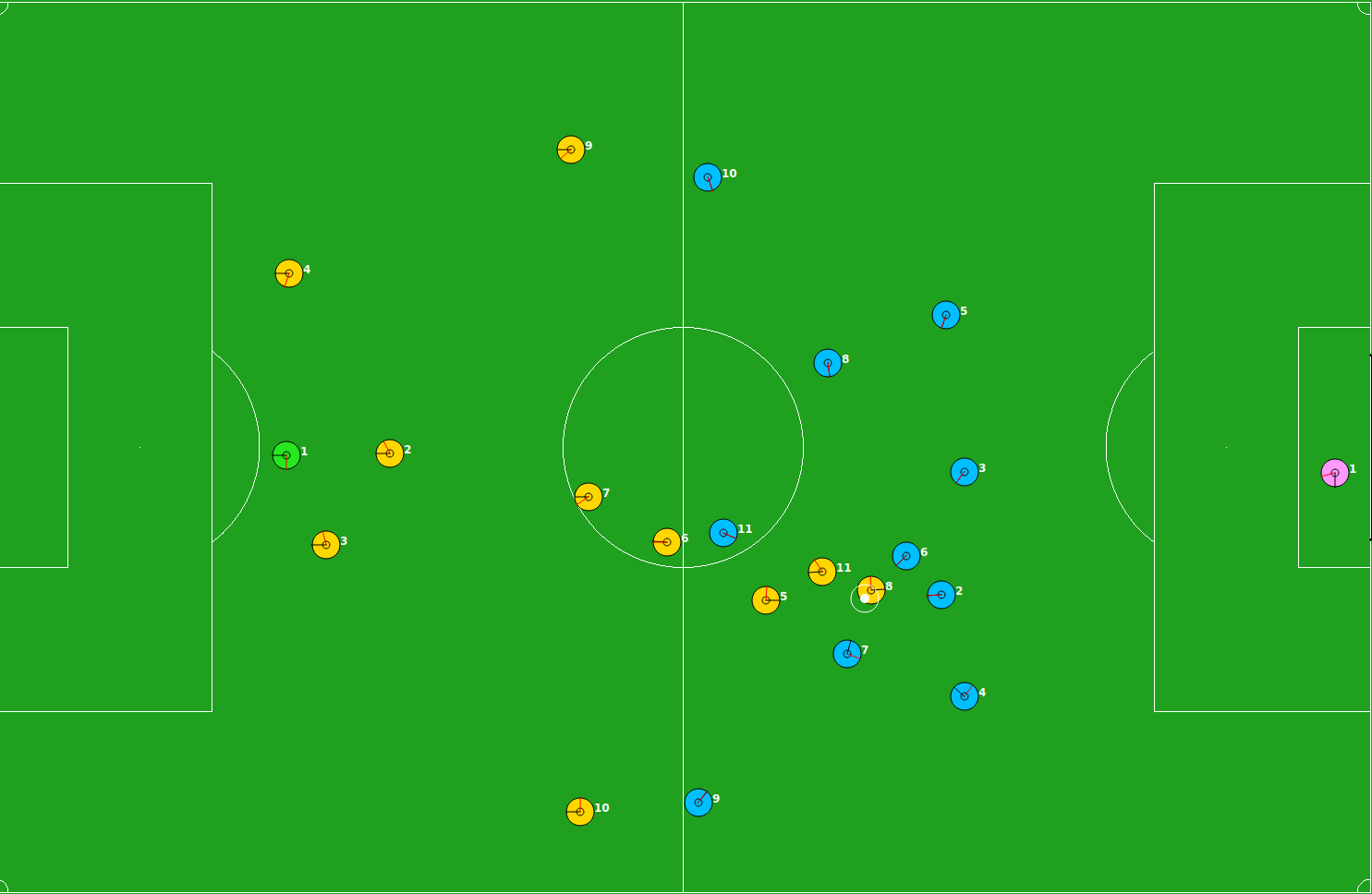}
\captionof{figure}{Qualitative off-ball movement example in the 11-vs-11 full-field benchmark. Yellow players are controlled left-team agents, blue players are {HELIOS-base-controlled opponents}, and the white circle indicates the ball. The visualization shows that, except for players close to the ball, some off-ball players far from the ball continue moving in fixed directions and do not yet form a coherent team shape. {This example is from an earlier 11-vs-11 evaluation.}}
\label{fig:full_field_offball_move}
\end{center}

\end{document}